\documentclass[10pt,twocolumn,letterpaper]{article}

\usepackage{iccv}
\usepackage{times}
\usepackage{epsfig}
\usepackage{graphicx}
\usepackage{subcaption}

\usepackage{amsfonts}
\usepackage{amsmath}
\usepackage{amssymb}
\usepackage{amsthm}
\usepackage{booktabs}

\usepackage{algorithm}
\usepackage{algpseudocode}

 \usepackage{relsize}
 
\usepackage{gensymb}
\usepackage[utf8]{inputenc}

\usepackage[pagebackref=true,breaklinks=true,letterpaper=true,colorlinks,bookmarks=false]{hyperref}
\iccvfinalcopy
\def\iccvPaperID{249}

\newcommand{\Darko}{$\textsc{Darko}$}
\newcommand{\Darkosp}{$\textsc{Darko}$\phantom{ }}
\newcommand{\fbar}{\bar{f}}
\newcommand{\fhat}{\hat{f}}

\usepackage{soul}

\newif\ifbody
\newif\ifbuildappendix
\bodytrue
\buildappendixtrue

\title{First-Person Activity Forecasting with Online Inverse Reinforcement Learning}
\author{Nicholas Rhinehart and Kris M. Kitani \\ Robotics Institute, Carnegie Mellon University \\ Pittsburgh, PA 15213 \\
{\tt \small \{nrhineha,kkitani\}@cs.cmu.edu}}
\date{November 2016}

\begin{document}
\maketitle

\begin{abstract}
    We address the problem of incrementally modeling and forecasting long-term goals of a first-person camera wearer: what the user will do, where they will go, and what goal they seek. In contrast to prior work in trajectory forecasting, our algorithm, \Darko, goes further to reason about semantic states (will I pick up an object?), and future goal states that are far in terms of both space and time. \Darkosp learns and forecasts from first-person visual observations of the user's daily behaviors via an Online Inverse Reinforcement Learning (IRL) approach. Classical IRL discovers only the rewards in a batch setting, whereas \Darkosp discovers the states, transitions, rewards, and goals of a user from streaming data. Among other results, we show \Darkosp forecasts goals better than competing methods in both noisy and ideal settings, and our approach is theoretically and empirically no-regret.
\end{abstract}

\ifbody

\section{Introduction} \label{sec:introduction}

Our long-term aim is to develop an AI system that can learn about a person's intent and goals by continuously observing their behavior. In this work, we progress towards this aim by proposing an online Inverse Reinforcement Learning (IRL) technique to learn a decision-theoretic human activity model from video captured by a wearable camera. The use of a wearable camera is critical to our task, as human activities must be observed up close and across large environments. Imagine a person's daily activities---perhaps they are at home today, moving about, completing tasks. Perhaps they are a scientist that conducts a long series of experiments across various stations in a laboratory, or they work in an office building where they walk about their floor, get coffee, \etc. As people tend to be very mobile, a wearable camera is ideal for observing a person's behavior.

\begin{figure}[th!]
\centering
\includegraphics[width=.5\textwidth]{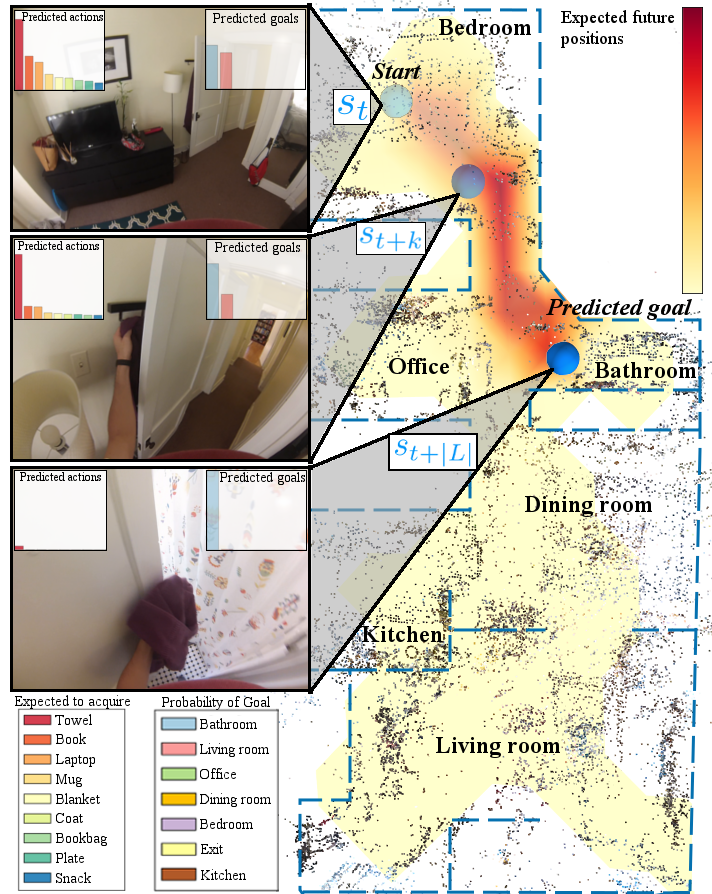}
\caption{\textbf{Forecasting future behavior from first-person video.} Overhead map shows likely future goal states. $s_i$ is user \textit{state} at time $i$. Histogram insets display predictions of user's long-term semantic goal (inner right) and acquired objects (inner left).}\vspace{-3mm}
\end{figure}

\begin{figure*}[t]
\begin{subfigure}[c]{.45\textwidth}
\centering
\includegraphics[height=1.4in]{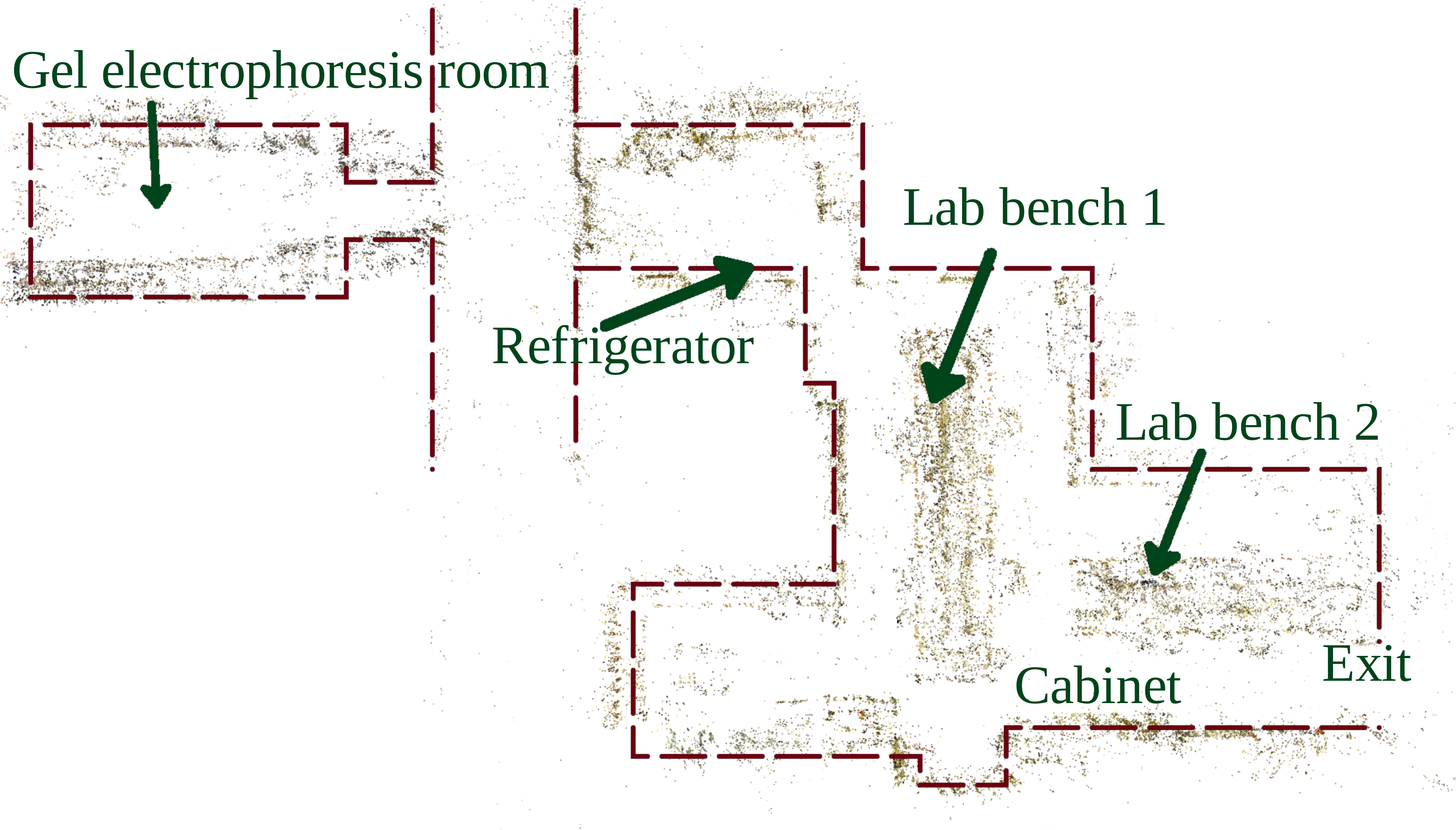}
\caption{Lab environment} \label{subfig:lab_overhead}
\end{subfigure}
\begin{subfigure}[c]{.55\textwidth}
\centering
\includegraphics[height=1.4in]{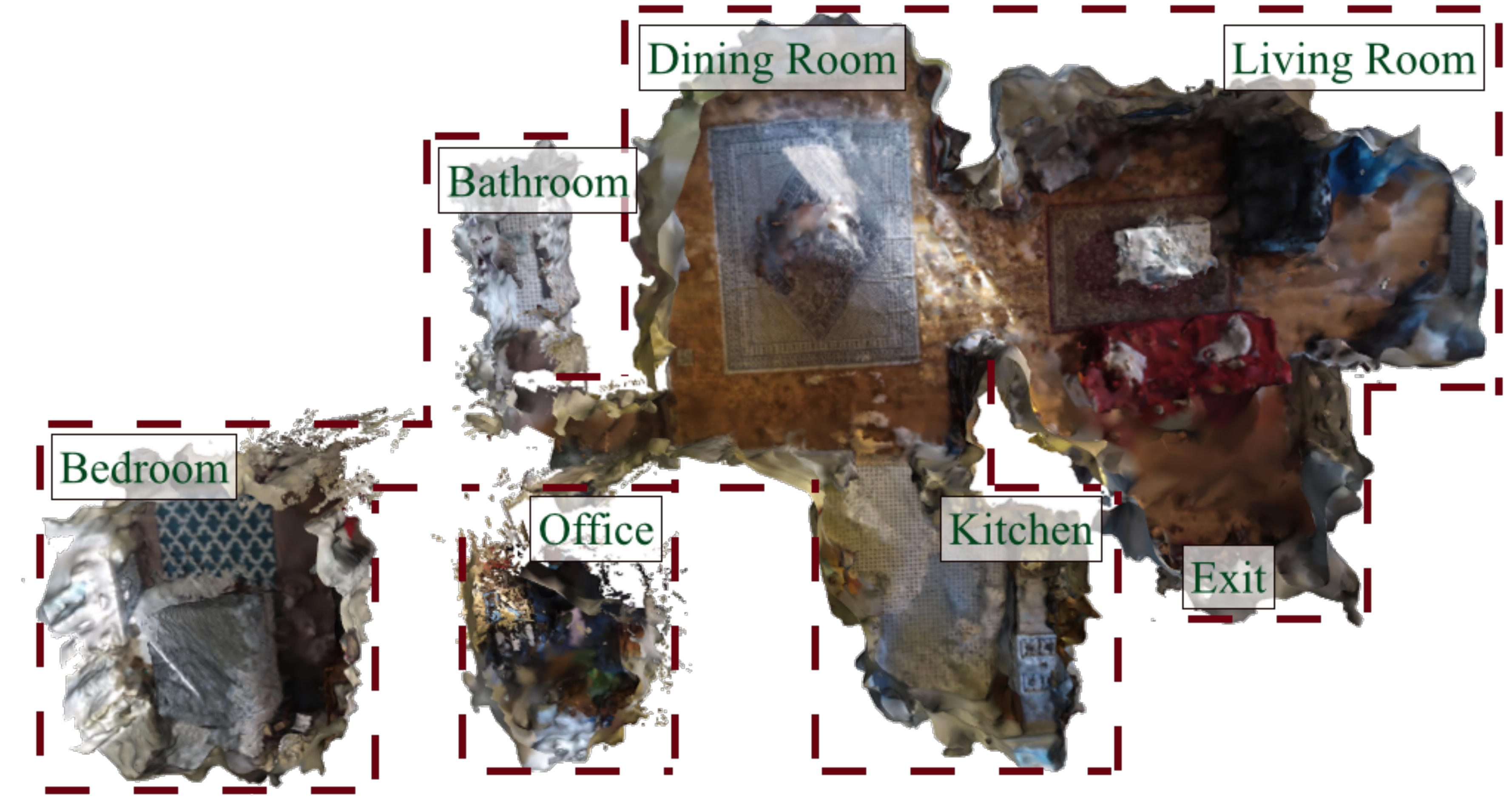}
\caption{Home 1 environment} \label{subfig:home_overhead}
\end{subfigure}
\vspace{-1mm}
\caption{Sparse SLAM points (\ref{subfig:lab_overhead}) and offline dense reconstruction (\ref{subfig:home_overhead}) using \cite{Furu:2010} for two of our dataset environments.} \label{fig:overhead_scenes}\vspace{-3mm}
\end{figure*}

Since our task is to continuously learn human behavior models (\ie., a policy) from observed behavior captured with a wearable camera, our task is best described as an online IRL problem. The problem is an \textit{inverse} Reinforcment Learning problem because the underlying reward or cost function of the person is unknown. We must infer it along with the policy from the demonstrated behaviors. Our task is also an \textit{online learning problem}, because our algorithm must continuously learn as a part of a life-long process. From this perspective, we must develop an online learning approach that learns effectively over time.

We present an algorithm that \textit{incrementally learns spatial and semantic intentions} (where you will go and what you will do) of a first-person camera wearer. By tracking the goals a person achieves, the algorithm builds a set of possible futures. At any time, the user's future is predicted among this set of goals. We term our algorithm ``Discovering Agent Rewards for K-futures Online" (\Darko), as it learns to associate rewards with semantic states and actions from demonstrations to predict among $K$ possible goals.

\noindent\textbf{Contributions:} To the best of our knowledge, we present the first application of ideas from online learning theory and inverse reinforcement learning to the task of continuously learning human behavior models with a wearable camera. Our proposed algorithm is distinct from traditional IRL problems as we jointly discover states, transitions, goals, and the reward function of the underlying Markov Decision Process model. Our proposed human behavior model also goes beyond first-person trajectory forecasting and allows for the prediction of future human activities that can happen outside the immediate field of view and far into the future.

\section{Related Work} \label{sec:related_work}

\noindent\textbf{First-person vision (FPV):} Wearable cameras have been used for various human behavior understanding tasks \cite{fathi2011understanding,lee2012discovering,Li_2015_CVPR,pirsiavash2012detecting,ryoo2013first} because they give direct access to detailed visual information about a person's actions. Leveraging this feature of FPV, recent work has shown that it is possible to predict where people will look during actions \cite{Li_2015_CVPR} and how people will use the environment \cite{rhinehart2016learning}.

\noindent\textbf{Decision-Theoretic Modeling:} Given agent demonstrations, the task of \textit{inverse reinforcement learning} (IRL) is to recover a \textit{reward function} of an underlying Markov Decision Process (MDP) \cite{abbeel2004apprenticeship}. IRL has been used to model taxi driver behavior \cite{ziebart2008maximum} and pedestrian behavior \cite{ziebart2009planning,kitani2012activity}. In contrast to previous work, we go beyond physical trajectory forecasting by reasoning over future object interactions and predicting future goals in terms of scene types. 

\noindent\textbf{Online Learning Theory:} The theory of learning to making optimal predictions from streaming data is well studied \cite{shalev2012online} but is rarely used in computer vision, compared to the more prevalent application of supervised learning theory. We believe, however, that the utility of online learning theory is likely to increase as the amount of available data for processing is ever increasing. While the concept of online learning has been applied to inverse reinforcement learning \cite{ratliff2006maximum}, the work was primarily theoretic in nature and has found limited application.

\noindent\textbf{Trajectory forecasting:} Physical trajectory forecasting has received much attention from the vision community. Multiple human trajectory forecasting from a surveillance camera was investigated by \cite{ma2016game}. Other trajectory forecasting approaches use demonstrations observed from a bird's-eye view; \cite{xie2013inferring} infers latent goal locations and \cite{Alahi_2016_CVPR} employ LSTMs to jointly reason about trajectories of multiple humans. In \cite{Park_2016_CVPR}, the model forecasted short-term future trajectories of a first-person camera wearer by retrieving the nearest neighbors from a dataset of first-person trajectories under an obstacle-avoidance cost function, with each trajectory representing predictions of where the user will move in view of the frame; in \cite{su2016social}, a similar model with learned cost function is extended to multiple users.

\noindent\textbf{Predicting Future Behavior:} In \cite{hoai2014max, ryoo2011human}, the tasks are to recognize an unfinished event or activity. In \cite{hoai2014max}, the model predicts the onset for a single facial action primitive, \eg the completion of a smile, which may take less than a second. Similarly, \cite{ryoo2011human} predicts the completion of short human to human interactions. In \cite{lan2014hierarchical}, a hierarchical structured SVM is employed to forecast actions about a second in the future, and \cite{Vondrick_2016_CVPR} demonstrates a semi-supervised approach for forecasting human actions a second into the future. Other works predict actions several seconds into the future \cite{Cao_2013_CVPR,koppula2016anticipating,li2014prediction,walker2014patch}. In contrast, we focus on high-level transitions over a sequence of future actions that may occur outside the frame of view, and take a longer time to complete (in our dataset, the mean time to completion is $21.4$ seconds).

\section{Online IRL with DARKO} \label{sec:approach}

Our goal is to forecast the future behaviors of a person from a continuous stream of video captured by a wearable camera. Given a continuous stream of FPV video, our approach extracts a sequence of state variables $\{ s_1, s_2,\dots \}$ using a portfolio of visual sensing algorithms (\eg., SLAM, stop detection, scene classification, action and object recognition). In an online fashion, we segment this state sequence into episodes (short trajectories) by discovering terminal goal states (\eg., when a person stops). Using the most recent episode, we adaptively solve the inverse reinforcement learning problem using online updates. Solving the IRL problem in an online fashion means that we incrementally learn the underlying decision process model.

\subsection{First-Person Behavior Model} \label{sec:terminology}

A Markov Decision Process (MDP) is commonly used to model the sequential decision process of a rational agent. In our case, we use it to describe the activity of a person with a wearable camera. In a typical reinforcement learning problem, all elements of the MDP are assumed to be known and the task is to estimate an optimal policy $\pi (a | s)$, that maps a state $s$ to an action $a$, by observing rewards. In our novel online inverse formulation, every element of the MDP, including the policy, is unknown and must be inferred as new video data arrives. Formally, an MDP is defined as:
\begin{align*}
\mathcal{M} = \{\mathcal{S}, \mathcal{A}, T(\cdot, \cdot), R_{\theta}(\cdot, \cdot)\}.
\end{align*}

\noindent\textbf{States:} $\mathcal{S}$ is the state space: the set of states an agent can visit. In our online formulation, $\mathcal{S}$ is initially empty, and must be expanded as new states are discovered. We define a state $s$ as a vector that includes the location of the person (3D position), the last place the person stopped (a previous goal state), and information about any object that the person might be holding. Formally, a state $s \in \mathcal{S}$ is denoted as: 
\begin{align*}
    s &= [x, y, z, o_1 \dots, o_{|\mathcal{O}|},  h_1, \dots h_{|\mathcal{K}|}].
\end{align*}

The triplet $[x, y, z]$ is a discrete 3D position. To obtain the position, we use a monocular visual \textsc{Slam} algorithm \cite{mur2015orb} to localize the agent in a continuously built map.

The vector $o_1 \dots, o_{|\mathcal{O}|}$ encodes any objects that the person is currently holding. We include this information in the state vector because the objects a user acquires are strongly correlated to the intended activity \cite{fathi2011understanding}. $o_j = 1$ if the user has object $j$ in their possession and zero otherwise. $\mathcal{O}$ is a set of pre-defined objects available to the user. $\mathcal{K}$ is a set of pre-defined scene types available to the user, which can be larger than the true number of scene types. The vector $h_1, \dots h_{K}$ encodes the last scene type the person stopped. Example scene types are { \small\texttt{kitchen}} and {\small\texttt{office}}. $h_i = 1$ if the user last arrived at scene type $i$ and is zero otherwise.

\noindent{\textbf{Goals:}} We also define a special type of state called a \textit{goal state} $s \subset \mathcal{S}_g$, to denote states where the person has achieved a goal. We assume that when a person stops, their location in the environment is a goal. We detect goal states by using a velocity-based stop detector. Whenever a goal state is encountered, the sequence of states since the last goal state to the current goal state is considered a completed episode $\xi$. The set of goals states $\mathcal{S}_g \subset \mathcal{S}$ expands with each detection. We explain later how $\mathcal{S}_g$ is used to perform goal forecasting.

\noindent\textbf{Actions:} $\mathcal{A}$ is the set of actions. $\mathcal{A}$ can be decomposed into two parts: $\mathcal{A} = \mathcal{A}_m \cup \mathcal{A}_c$. The act of moving from one location in the environment to another location is denoted as $a_m \in \mathcal{A}_m$. Like $\mathcal{S}$, $\mathcal{A}_m$ must be built incrementally. The set $\mathcal{A}_c$ is the set of possible {\small \texttt{acquire}} and {\small \texttt{release}} actions of each object: $\mathcal{A}_c = \{{\small \texttt{acquire}, \texttt{release}}\} \times \mathcal{O}$. The act of releasing or picking up an object is denoted as $a_c \in \mathcal{A}_c$. Each action $a_c$ must be detected. We do so with an image-based first-person action classifier. More complex approaches could improve performance \cite{ma2016going}.

\noindent\textbf{Transition Function:} The transition function $T: (s,a) \mapsto s'$ represents how actions move a person from one state to the next state. $T$ is constructed incrementally as new states are observed and new actions are performed. In our work, $T$ is built by keeping a table of observed $(s,a,s')$ triplets, which describes the connectivity graph over the state space. More advanced methods could also be used to infer more complex transition dynamics \cite{sun2016learning,wei1994time}. 

\noindent\textbf{Reward Function:} $R(s,a;\theta)$ is an instantaneous reward function of action $a$ at state $s$. We model $R$ as the inner product between a vector of features $f(s, a)$ and a vector of weights $\theta$. The reward function is essential in value-based reinforcement learning methods (in contrast to policy search methods) as it is used to compute the policy $\pi(a|s)$. In the maximum entropy setting, the policy is given by $\pi(a | s) \propto e^{Q(s,a) - V(s)}$, where the value functions $V(s)$ and $Q(s,a)$ are computed from the reward function by solving the Bellman equations \cite{ziebart2008maximum}. In our context, we learn the reward function online.

Intuitively, we would like the features $f$ of the reward function to incorporate information such as the position in an environment or objects in possession, since it is reasonable to believe that the goal of many activities is to reach a certain room or to retrieve a certain object. To this end, we define the features of the reward to mirror the information already contained in the state $s_t$: the position, previous scene type, and objects held. To be concrete, the feature vector $f(s,a)$ is the concatenation of the 3-d position coordinates $[x,y,z]$, a $K$-dimensional indicator vector over previous goal state type and a $\mathcal{|O|}$-dimensional indicator vector over held objects. We also concatenate a $|\mathcal{A}_c|$-dimensional indicator vector over actions $a_c \in \mathcal{A}_c$.

\subsection{The DARKO Algorithm} \label{sec:darko_algorithm}

We now describe our proposed algorithm for incrementally learning all MDP parameters, most importantly the reward function, given a continuous stream of first-person video (see \Darkosp in Algorithm~\ref{alg:darko}). The procedure begins by initializing $s$, reward parameters $\theta$, empty state space $\mathcal{S}$, goal space $\mathcal{S}_g$, transition function $T$, and current episode $\xi$. 

\noindent\textbf{State Space Update}: Image frames are obtained from a first-person camera (the $\textsc{NewFrame}$ function), and $\textsc{Slam}$ is used to track the user's location (lines~\ref{algline:newframe} and \ref{algline:slam}). An image-based action detection algorithm, $\textsc{ActDet}$, detects hand-object interactions $a_c$ and decides movements $a_m$ as a function of current and previous position. While we provide an effective method for $\textsc{ActDet}$, our focus is to integrate (rather than optimize) its outputs. Lines~\ref{algline:trajgrow} and \ref{algline:transition} show how the trajectory is updated and MDP parameters of state space and transition function are expanded. Line~\ref{algline:forecast} represents a collection of generalized forecasting tasks (see Section \ref{sec:future_goal_prediction}), such as the computation of future goal posterior and trajectory forecasting. 

\noindent\textbf{Goal Detection}: In order to discover goals, we use a stop-detection algorithm $\textsc{GoalDet}$, by using the camera velocity computed from SLAM (Line~\ref{algline:goaldet}). If a goal state has been detected, that terminal state is added to the set of goal states $\mathcal{S}_g$. The detection of a terminal state also marks the end of an episode $\xi$. The previous goal state type is also updated for the next episode. Again, while we provide an effective method for $\textsc{GoalDet}$, our focus is to integrate (rather than optimize) its outputs. 

\noindent\textbf{Online IRL}: With the termination of each episode $\xi$, the reward function $R$ and corresponding policy $\pi$ are updated via the reward parameters $\theta$ (Line~\ref{algline:fit}). The parameter update uses a sequence of demonstrated behavior via the episode $\xi$, and the current parameters of the MDP. More specifically, $\textsc{OnlineIRL}$ (Algorithm~\ref{alg:policy_fit}) performs online gradient descent on the likelihood under the maximum entropy distribution by updating current parameters of the reward function. The gradient of the loss can be shown to be the difference between expected feature counts $\bar{f}$ and empirical feature counts $\hat{f}$ of the current episode. Computing the gradient requires solving the soft value iteration algorithm of \cite{ziebart2010modeling}. We include a projection step to ensure $\|\theta\|_2 \leq B$.

To the best of our knowledge, this is the first work to propose an online algorithm for maximum entropy IRL in the streaming data setting. Following the standard procedure for ensuring good performance of an online algorithm, we analyze our algorithm in terms of the regret bound. The regret $\mathcal{R}_t$ of any online algorithm is defined as:
\begin{align*}
    \mathcal{R}_t = \sum_{i=0}^t l_t(\xi_t; \theta_t) - \min_{\theta^*} \sum_{i=0}^t l(\xi_t; \theta^*).
\end{align*}
The regret is the cumulative difference between the performance of the online model using current parameter $\theta$ versus the best hindsight model using the best parameters $\theta^{*}$. The loss $l_t$ is a function of the $t$'th demonstrated trajectory, and measures how well the model explains the trajectory.

In our setup, the loss function is defined as $l_t(\xi_t; \theta)= -\frac{1}{|\xi_t|} \sum_{i=0}^{|\xi_t|}\log{\pi_\theta(a_i | s_i)}$. It can be shown\footnote{Proof of the regret bound is given in the Appendix} that the regret of our online algorithm is bounded as follows:
\begin{align}
    \mathcal{R}_t \leq 2B\sqrt{2td}, \label{eq:regret}
\end{align}
where $B$ is a norm bound on $\theta$, $d$ is feature dimensionality, and $t$ is the episode. The average regret $\frac{\mathcal{R}_t}{t}$ approaches zero as $t$ grows since the bound is sub-linear in $t$. Verification of this no-regret property is shown in the experiments.

\algrenewcommand\algorithmicfunction{\textbf{procedure}}

\begin{algorithm}[tb]
\caption{\Darko: Discovering Agent Rewards for K-futures Online} \label{alg:darko}
\begin{algorithmic}[1]
\Function{\Darko}{\textsc{Slam}, \textsc{ActDet}, \textsc{GoalDet}}
\State $s \gets \mathbf{0}, \theta = \mathbf{0}, \mathcal{S} = \left\{\right\}, \mathcal{S}_g = \left\{\right\}$, \Call{T.Init}{\null}, $\xi = []$
\While{$\text{True}$}
\State frame $\gets$ \Call{NewFrame}{\null} \label{algline:newframe}
\State  $[x, y, z] \gets$ \Call{Slam.Track}{frame}  \label{algline:slam}
\State $a\gets$ \Call{ActDet}{$[x,y,z]$, frame}
\State $\xi \gets \xi \oplus (s, a), \mathcal{S} \gets \mathcal{S} \cup \left\{s\right\}$ \label{algline:trajgrow}
\State \Call{T.Expand}{$s$, $a$}, $s \gets T(s, a)$ \label{algline:transition}
\State $\blacktriangleright~\text{Goal forecasting, trajectory forecasting}, \ldots$ \label{algline:forecast}
\State is\_goal $\gets$ \textsc{GoalDet}($s$, frame, $\mathcal{S}_g$) \label{algline:goaldet}
\If{is\_goal}
    \State $\mathcal{S}_g \gets \mathcal{S}_g \cup \left\{s\right\}$
    \State $\pi,\theta \gets$ \Call{OnlineIRL}{$\theta$, $\mathcal{S}$, $T$, $\xi$, $\mathcal{S}_g$} \label{algline:fit}
    \State $s \gets T(s, a = \text{at\_goal}), \xi = []$
\EndIf
\EndWhile
\EndFunction
\end{algorithmic}
\end{algorithm}

\begin{algorithm}[tb]
\caption{Online Inverse Reinforcement Learning} \label{alg:policy_fit}
\begin{algorithmic}[1]
\Function{OnlineIRL}{$\theta$, $\mathcal{S}$, $T$, $\xi$, $\mathcal{S}_g$; $\lambda$, $B$}
     \State $\overline{f}_i = \sum_{(s,a)\in \xi}f(s, a)$
     \State $\blacktriangleright \text{Compute}~R(s,a;\theta)~\forall s \in \mathcal{S}, a \in \mathcal{A}$
     \State $\pi \gets$ \Call{SoftValueIteration}{$R$, $\mathcal{S}, \mathcal{S}_g, T$}
    \State $\hat{f}_i \gets E_{\pi}\left[f(s,a)\right]$
    \State $\theta \gets \text{proj}_{\|\theta\|_2 \leq B}(\theta - \lambda (\overline{f}_i - \hat{f}_i))$
\State \Return $\pi, \theta$
\EndFunction
\end{algorithmic}
\end{algorithm}

\algrenewcommand\algorithmicfunction{\textbf{function}}

\section{Generalized Activity Forecasting}

Without Line~\ref{algline:forecast}, Algorithm~\ref{alg:darko} only describes our online IRL process to infer the reward function. In order to make incremental predictions about the person's future behaviors online, we can leverage the current MDP and reward function. An important function which lays the basis for predicting future behaviors is the state visitation function, denoted $D$. We now show how $D$ can be modified to perform generalized queries about future behavior.

\subsection{State Visitation Function $D$} \label{sec:future_state_and_subspace_prediction}

Using the current estimate of the MDP and the reward function, we can compute the policy of the agent. Using the policy, we can forward simulate a distribution of all possible futures. This distribution is called the state visitation distribution \cite{ziebart2010modeling}. More formally, the posterior expected count of future visitation to a state $s_x$ can be defined as
\begin{align} \label{eq:expected_count_visitation}
D_{s_x | \xi_{0 \rightarrow t}} &\triangleq \mathbb{E}_{P(\xi_{t+1 \rightarrow T} | \xi_{0 \rightarrow t})}\left[\sum_{\tau = t+1}^T I(s_{\tau} = s_x) \right].
\end{align} This quantity represents the agent's expectation to visit each state in the future given the partial trajectory. $\xi_{0 \rightarrow t}$ indicates a partial trajectory starting at time $0$ and ending at time $t$. The expectation is taken under the maximum causal entropy distribution, $P(\xi_{t+1 \rightarrow T} | \xi_{0 \rightarrow t})$, which gives the probability of a future trajectory given the current trajectory. $I$ is the indicator function, which counts agent visits to $s_x$. Equation~\ref{eq:expected_count_visitation} is estimated by sampling trajectories from $\pi_\theta(a|s)$.

\subsection{Activity Forecasting with State Subsets}

In this work, we extend the idea of state visitations to a single state $s_x$ to a more general \textit{subset of states} $\mathcal{S}_p$. While a generalized prediction task was not particularly meaningful in the context of trajectory prediction \cite{kitani2012activity,ziebart2008maximum}, predictions over a subset of states now represents semantically meaningful concepts in our proposed MDP. By using the state space representation of our first-person behavior model, we can construct subsets of the state space that have interesting semantic meaning, such as ``having an object $o_i$" or ``all states closest to goal $k$ with $\mathcal{O}_j$ set of objects." 

Formally, we define the expected count of visitation to a subset of states $\mathcal{S}_p$ satisfying some property $p$:
\begin{align} \label{eq:expected_subset_visitation}
D_{\mathcal{S}_p | \xi_{0 \rightarrow t}} &\triangleq \mathbb{E}_{P(\xi_{t+1 \rightarrow T} | \xi_{0 \rightarrow t})}\left[\sum_{\tau = t+1}^T I(s_{\tau} \in \mathcal{S}_p) \right] \\
&= \sum_{s_x \in \mathcal{S}_p}\mathbb{E}_{P(\xi_{t+1 \rightarrow T} | \xi_{0 \rightarrow t})}\left[\sum_{\tau = t+1}^T I(s_{\tau} = s_x) \right] \nonumber \\
&= \sum_{s_x \in \mathcal{S}_p} D_{s_x | \xi_{0 \rightarrow t}}. \label{eq:exp_count_result}
\end{align}
Equation \ref{eq:exp_count_result} is essentially marginalizing over the state subspace of Equation~\ref{eq:expected_count_visitation}. Derivation of two other inference tasks is given in the Appendix: (1) expected visitation prediction of performing an action after arriving at a subspace, (2) expected joint action-state subspace visitation prediction.

\subsection{Forecasting Trajectory Length}
In the following, we present a method to predict the length of the future trajectory. Formally, we can denote the expected trajectory length:
    \begin{align}
    \hat{\tau}_{\xi_{t+1 \rightarrow T} | \xi_{0 \rightarrow t}} &\triangleq \mathbb{E}_{P(\xi_{t+1 \rightarrow T} | \xi_{0 \rightarrow t})}\left|\xi_{t+1 \rightarrow T}\right| \label{eq:exp_traj_len}
    \end{align}
    Consider evaluating $D_{\mathcal{S}_p | \xi_{0 \rightarrow t}}$ from Equation~\ref{eq:exp_count_result} by setting $\mathcal{S}_p = \mathcal{S}$, that is, by considering the expected future visitation count to the entire state space. Then,
    \begin{align} 
D_{\mathcal{S} | \xi_{0 \rightarrow t}} &=\mathbb{E}_{P(\xi_{t+1 \rightarrow T} | \xi_{0 \rightarrow t})}\left[\sum_{\tau = t+1}^T I(s_{\tau} \in \mathcal{S}) \right] \nonumber \\
 &=\mathbb{E}_{P(\xi_{t+1 \rightarrow T} | \xi_{0 \rightarrow t})}\left[\sum_{\tau = t+1}^T 1 \right] \nonumber \\
&= \mathbb{E}\left |\xi_{t+1 \rightarrow T}\right | = \hat{\tau}_{\xi_{t+1 \rightarrow T} | \xi_{0 \rightarrow t}} \label{eq:length}
\end{align}
where $\left |\xi \right |$ indicates the number of states in trajectory $\xi$.

\subsection{Future Goal Prediction} \label{sec:future_goal_prediction}

As previously described, we wish to predict the final goal of a person's action sequence. For example, if I went to the study to pick up a cup, how likely am I to go to the kitchen versus the living room? This problem can be posed as solving for the MAP estimate of $P(g | \xi) \forall g \in \mathcal{S}_g$, the posterior over goals. It describes \textit{what goal the user seeks given their current trajectory}, defined as:
\begin{align} \label{eq:posterior_deterministic}
    P(g | \xi_{0\rightarrow t}) \propto P(g)e^{V_{s_t}(g) - V_{s_0}(g)}, 
\end{align} where $V_{s_i}(g)$ is the value of $g$ with respect to a partial trajectory that ends in $s_i$. Notice that the likelihood term is exponentially proportional to the value difference between the start state $s_0$ and the current state $s_t$. In this way, the likelihood encodes the progress made towards a goal $g$ in terms of the value function.

\section{Experiments}

We first present the dataset we collected. Then, we discuss our methods for goal discovery and action recognition. To reiterate, our focus is not to engineer these methods, but instead to make intelligent use of their outputs in \Darkosp for the purpose of behavior modeling. We compare \Darko's performance versus several baselines on the task of goal forecasting, and show \Darko's performance is superior. Then, we analyze \Darko's performance under less noisy conditions, to illustrate how it improves when provided with more robust goal discovery and action detection algorithms. Next, we illustrate \Darkosp's empirical no-regret performance, which further shows it is an efficient online learning algorithm. Finally, we present trajectory length forecasting results, and find that our length forecasts exhibit low median error. Additional analyses including feature ablation and incorporating uncertainty from goal discovery are presented in the Appendix.

\subsection{First-Person Continuous Activity Dataset} \label{sec:dataset}

We collected a dataset of sequential goal-seeking behavior in several different environments such as home, office and laboratory. The users recorded a series of activities that naturally occur in each scenario. Each user helped design the script they followed, which involved their prior assumptions about what objects they will use and what goal they will seek. An example direction from a script is ``obtain a snack and plate in kitchen, eat at dining room table." 

Users wore a hat-mounted Go-Pro Hero camera with $94\degree$ vertical, $123\degree$ horizontal FOVs. Our dataset is comprised of 5 user environments, and includes over 250 actions with 19 objects, 17 different scene types, at least 6 activity goals per environment, and about 200 high-level activities (trajectories). In each environment, the user recorded 3--4 long sequences of high-level activities, where each sequence represents a full day of behavior. Our dataset represents over 15 days of recording. 

For evaluation, all ground truth labels of objects (\eg cup, backpack), actions (\ie acquire, release) and goals (\eg kitchen, bedroom) were first manually annotated. A goal label correspond to \textit{when} a high-level direction was completed, and \textit{in which scene} it was completed, \eg~(dining room, time=$65s$). An action label indicates when an activity was performed, \eg~(acquire, cup, time=$25s$). Further details are presented in the Appendix.

\begin{figure}[t]
\centering
\includegraphics[width=.99\linewidth]{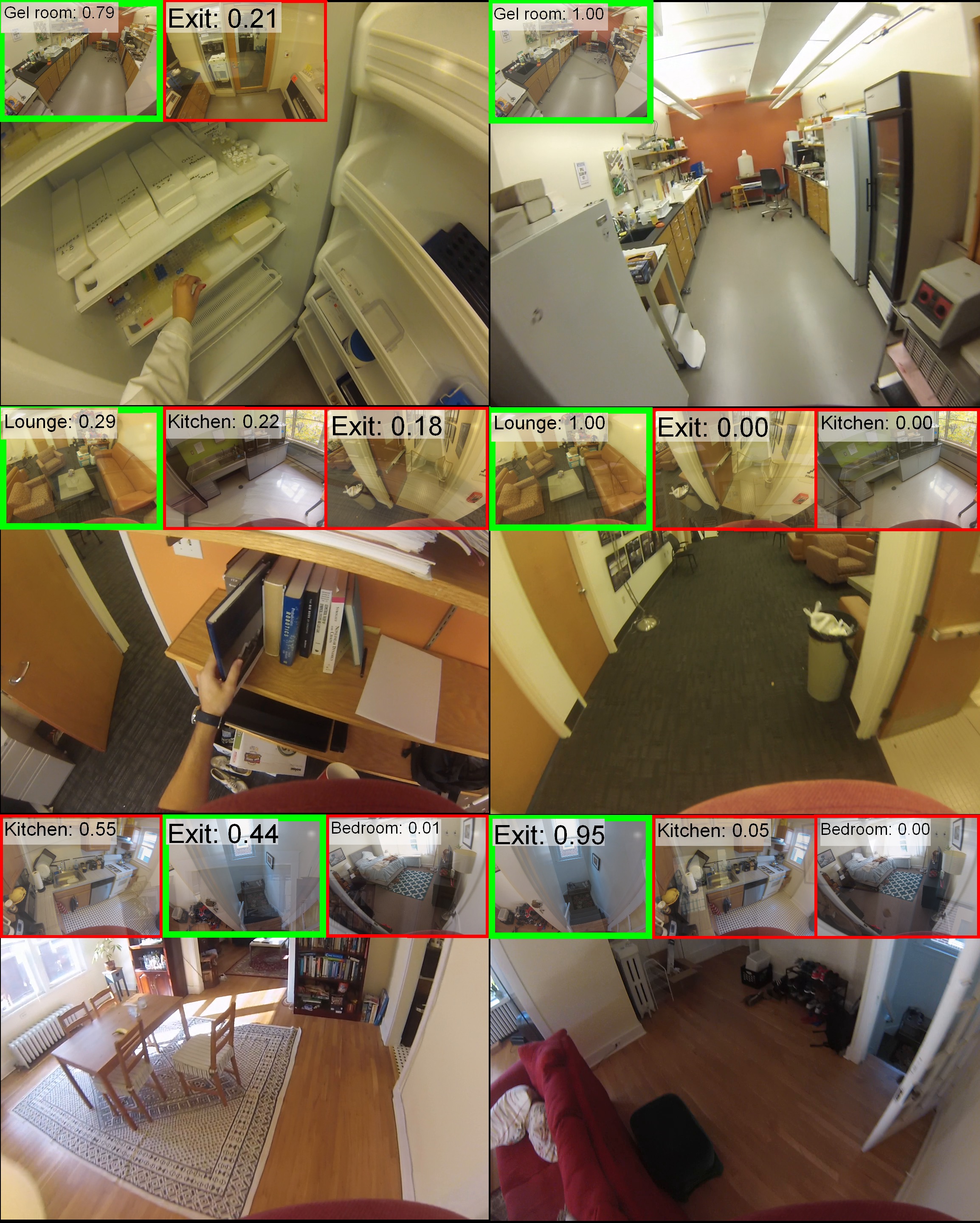}
\caption{\textbf{Goal forecasting examples:} A temporal sequence of goal forecasting results is shown in each row from left to right, with the forecasted goal icons and sorted goal probabilities inset (green: $P(g^*|\xi)$, red: $P(g_i\neq g^*|\xi)$).\\ \textit{Top}: the scientist acquires a sample to boil in the gel electrophoresis room. \textit{Middle}: the user gets a textbook and goes to the lounge. \textit{Bottom}: the user leaves their apartment.} \label{fig:posterior_image_visualization}
\end{figure}

\subsection{Goal Discovery and Action Recognition}
We describe two goal discovery methods and an action recognition method that can serve
as input to \Darko. With respect to Algorithm~\ref{alg:darko}, these are $\textsc{GoalDet}$ and $\textsc{ActDet}$.

\noindent\textbf{Scene-based Goal Discovery:} This model assumes that if a scene classifier is very confident in the scene type for several images frames, the camera wearer must be in a meaningful place in the environment (\ie., kitchen, bedroom, office). We use the output of a scene classifier from \cite{zhou2014learning} (GoogLeNet model) on every frame from the wearable camera. If the mean scene classifier probability for a scene type is above a threshold $\rho_g$ for 20 consecutive image frames, then we add the current state $s_t$ to the set of goals $\mathcal{S}_g$.

\noindent\textbf{Stop-based Goal Discovery:} This model assumes that when a person stops, they are at an important location. Using \textsc{Slam}'s 3D camera positions, we apply a threshold on velocity to detect stops in motion. When a stop is detected, we add the current state $s_t$ to the set of goals $\mathcal{S}_g$. In Table~\ref{tab:goal_discovery_performance}, temporal accuracies are computed by counting detections within $3$-second windows of ground truth labels as true positives; for the scene-based method, a true positive also requires the scene type to match the ground truth scene type. Stop-based discovery is reliable across all environments, thus, we use it as our primary goal discovery method.

\begin{table}[tb]
\centering
\footnotesize
{\setlength{\tabcolsep}{4pt}
\begin{tabular}{@{}llllll@{}}
   \toprule
 Method  & Home 1 & Home 2 & Office 1 & Office 2 & Lab 1 \\
\midrule
 Scene Discovery &  $0.93$ & $0.24$ & $0.62$ & $0.49$ & $0.32$ \\
 Stop Discovery  & $0.62$ & $0.68$ & $0.67$ & $0.69$ & $0.73$ \\
\midrule
 Act. Recognition & $0.64$ &$0.63$ & $0.66$ & $0.56$ & $0.71$  \\
 \bottomrule
\end{tabular}
}
\caption{\textbf{Goal Discovery and Action Recognition.} The per-scene goal discovery and action recognition accuracies are shown for our methods. A $3$-second window is used around every goal discovery to compute accuracy.} \label{tab:goal_discovery_performance} 
\end{table}
\noindent\textbf{Image-based Object Recognition:} We designed an object recognition approach that classifies the object the user interacts with at every temporally-labeled window. It overwrites the ground-truth object label with its detection. The approach first detects regions of \texttt{\small person} in each frame with \cite{redmon2016you} to focus on objects near the visible hands, which are cropped with context and fed into an image-classifier trained on ImageNet \cite{simonyan2014very}. The outputs are remapped to our object set, and a final classification is produced by taking the maximum across objects. The per-action classification accuracies in Table~\ref{tab:goal_discovery_performance} demonstrate the method can produce reasonable action classifications across all scenes. While imperfect, these detections serve as useful input to \Darko.
\subsection{Goal Forecasting Performance} \label{sec:goal_posterior_forecasting}

At every time step, our method predicts the user's goal or final destination (\eg., bedroom, exit) as described in Section \ref{sec:future_goal_prediction} and shown in Figure~\ref{fig:posterior_image_visualization}. To understand the goal prediction reliability, we compare our approach to several baseline methods for estimating the goal posterior $P(g | \xi_{0\rightarrow t})$, where $g$ is a goal and $\xi_{0\rightarrow t}$ is the observed state sequence up to the current time step. Each baseline requires the state tracking and goal discovery components of \Darko.

\noindent\textbf{Uniform Model (Uniform):} This model returns a uniform posterior over possible goals $P_n(g) = 1/K_n$ known at the current episode $n$, defining worst case performance.

\noindent\textbf{Logistic Regression Model (Logistic):} A logistic regression model $P_n(g|s_t)$ is fit to map states $s_t$ to goals $g$.

\noindent\textbf{Max-Margin Event Detection (MMED) \cite{hoai2014max}:} A set of max-margin models $P_n(g| \phi(s_{t:t-w}))$ are trained to map features $\phi$ of a $w$-step history of state vectors $s_{t:t-w}$ to a goal score. We use the best performing \texttt{sumL1norm} features provided with the publicly available code.

\noindent\textbf{RNN Classifier (RNN):} An RNN is trained to predict $P_n(g|\xi_{0\rightarrow t})$. We experimented with a variety of parameters (see the Appendix) and report the best results. 

Since all methods above are online algorithms, each of the models $P_n$ is updated after every episode $n$. In order quantify performance with a single score, we use the mean probability assigned to the ground truth goal type $g^{*}$ over all episodes $\{\xi_n\}_{n=1}^{N}$ as $\overline{P}(g^{*} | \{\xi_n\}_{n=1}^{N}) = \frac{1}{N} \sum_{n=1}^{N} \sum_{t=1}^{T_n} P_n(g | \xi_{nt})$ (also denoted $\overline{P}_{g^{*}}$). The goal forecasting performance results are summarized in  Table~\ref{tab:aucs} using the above metric. 

\begin{table}[tb]
\centering
\footnotesize
\begin{tabular}{@{}llllll@{}}
\toprule
Method & Home 1 & Home 2 & Office 1 & Office 2 & Lab 1 \\
 \midrule
 \textbf{\Darko} & $\mathbf{0.524}$ & $\mathbf{0.378}$  & $\mathbf{0.667}$ & $\mathbf{0.392}$ & $\mathbf{0.473}$ \\
MMED \cite{hoai2014max} & $0.403$ & $0.299$ & $0.600$& $0.382$ & $0.297$ \\
RNN & $0.291$ & $0.274$ & $0.397$ & $0.313$ & $0.455$ \\
Logistic & $0.458$ & $0.297$ & $0.569$ & $0.323$ & $0.348$ \\
Uniform & $0.181$ & $0.098$ & $0.233$ & $0.111$ & $0.113$ \\
\bottomrule
\end{tabular}
\caption{\textbf{Goal Forecasting Results (Visual Detections):} Proposed goal posterior (Sec.\ref{sec:future_goal_prediction}) achieves best $\overline{P}_{g^{*}}$ (mean probability of true goal).} \label{tab:aucs} 
\end{table}

\subsection{Goal Forecasting with Perfect Visual Detectors}

The experimental results up to this point have exclusively used visual detectors as input (\eg., SLAM, scene classification, object recognition). While we have shown that our approach learns meaningful human activity models from real computer vision input, we would also like to understand how our online IRL method performs when decoupled from the noise of the vision-based input. We perform the same experiments described in Section \ref{sec:goal_posterior_forecasting} but with idealized (ground truth) inputs for goal discovery and action recognition. We still use \textsc{Slam} for localization.

Table \ref{tab:perfect} summarizes the mean true goal probability for each of the dataset environments. We observe a mean absolute performance improvement of $0.27$ by using idealized inputs. Our proposed model continues to perform the best against baselines methods. This performance indicates that as vision-based component technologies improve, we can expect significant improvements in the ability to predict the goals of human activities.

We also measure performance when the action detection is built from ground truth and the goal discovery is built from our described methods. Our expectation is that \Darkosp with stop-based discovery should outperform \Darkosp with scene-based based discovery, given the stop-detector's more reliable goal detection performance (Table~\ref{tab:goal_discovery_performance}). The results over the dataset are given in Table~\ref{tab:aucs_detection}, confirming our expectation.

\begin{table}[tb]
\centering
\footnotesize
\begin{tabular}{@{}llllll@{}}
 \toprule 
Method & Home 1 & Home 2 & Office 1 & Office 2 & Lab 1 \\
 \midrule
 \textbf{\Darko} & $\mathbf{0.851}$ & $\mathbf{0.683}$  & $\mathbf{0.700}$ & $\mathbf{0.666}$ & $\mathbf{0.880}$ \\
MMED \cite{hoai2014max} & $0.648$ & $0.563$ & $0.589$ & $0.624$ & $0.683$ \\
RNN & $0.441$ & $0.322$ & $0.504$ & $0.454$ & $0.651$ \\
Logistic & $0.517$ & $0.519$ & $0.650$ & $0.657$ & $0.774$ \\
Uniform & $0.153$ & $0.128$ & $0.154$ & $0.151$ & $0.167$ \\
\bottomrule
\end{tabular}
\caption{\textbf{Goal Forecasting Results (Labelled Detections):} Proposed goal posterior achieves best $\overline{P}_{g^{*}}$ (mean probability of true goal). Methods benefit from better detections.} \label{tab:perfect} 
\end{table}

\begin{table}[t]
\centering
\footnotesize
\begin{tabular}{@{}llllll@{}}
\toprule
 Method & Home 1 & Home 2 & Office 1 & Office 2 & Lab 1 \\
 \midrule
Scene-based  & $0.438$ & $0.346$ &$0.560$ & $0.238$  & $0.426$ \\ 
Stop-based & $0.614$ & $0.395$ & $0.644$ & $0.625$ & $0.709$ \\
\bottomrule
\end{tabular}
\caption{\textbf{Visual goal discovery:} Better goal discovery (cf. Table~\ref{tab:goal_discovery_performance}) yields better $\overline{P}_{g^{*}}$. Here, action detection labels are used to isolate performance differences.} \label{tab:aucs_detection}
\end{table}

\subsection{Goal Forecasting Performance over Time} 

In additional to understanding the performance of goal prediction with a single score, we also plot the performance of goal prediction over time. We evaluate the goal forecasting performance as a function of the fraction of time until reaching the goal. In Figure~\ref{fig:auc_over_time}, we plot the \textit{mean} probability of the true goal at each fractional time step $\widehat{P}(g^* | \xi_t) = \frac{1}{N} \sum_{n=1}^{N} P_n(g^* | \xi_{nt})$. Using fractional trajectory length allows for a performance comparison across trajectories of different lengths.

As shown in Figure~\ref{fig:auc_over_time}, \Darkosp exhibits the property of maintaining uncertainty early in the trajectory and converging to the correct prediction as time elapses in most cases. In contrast, the logistic regression, RNN, and MMED perform worse at most time steps. As it approaches the goal, our method always produces a higher confidence in the correct goal with lower variance. We tried \texttt{\small argmax} and Platt scaling \cite{platt1999probabilistic} to perform multi-class prediction with MMED; \texttt{\small argmax} yielded higher $\overline{P}_{g^{*}}$, in addition to making $\widehat{P}_{g^*}$ noisier. While the RNN sees many states, its trajectory-centric hidden-state representation may not have enough data to generalize as well as the state-centric baselines.

\begin{figure}[t]
\centering
\includegraphics[width=\columnwidth]{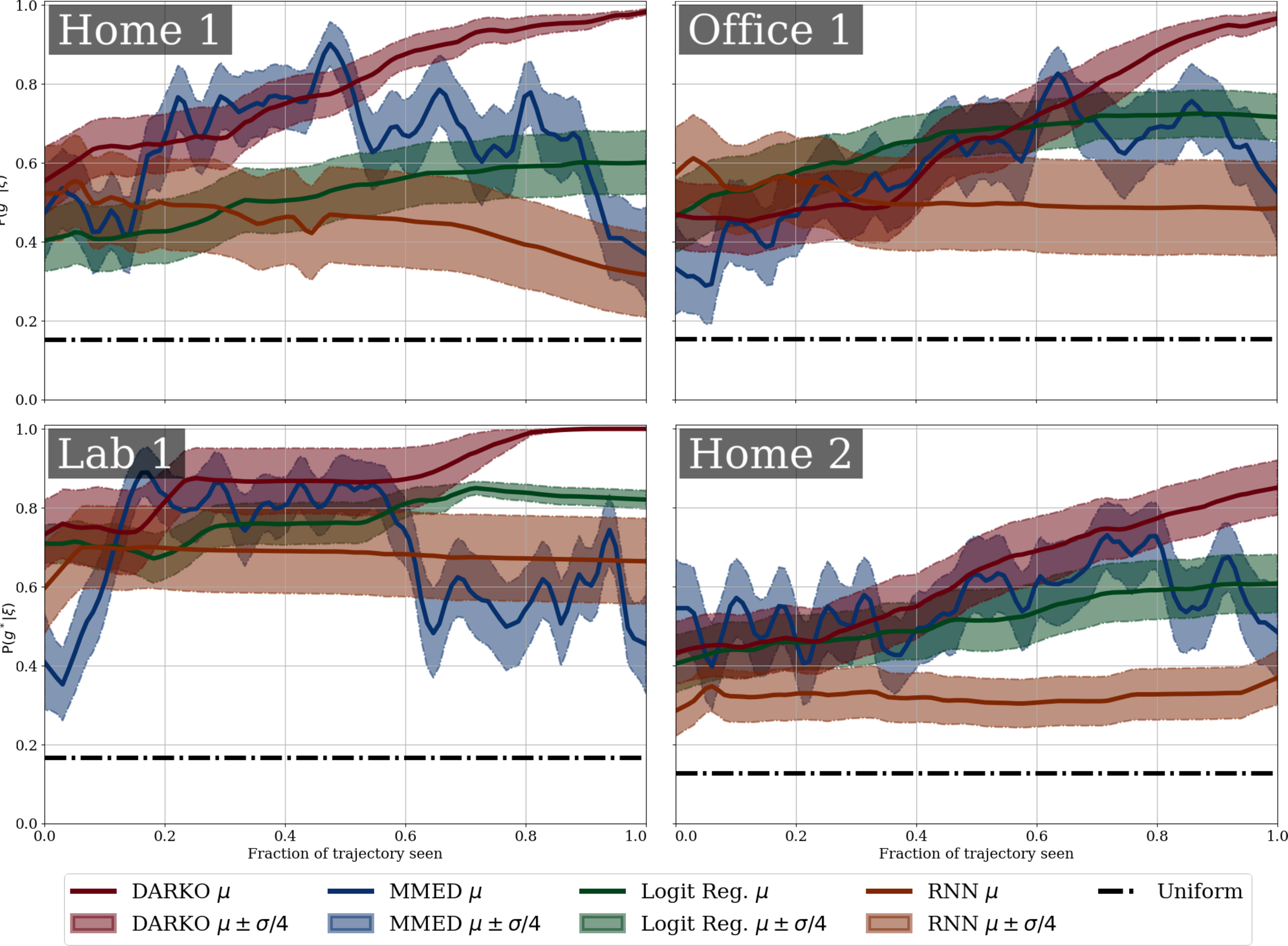}
\caption{\textbf{Goal posterior forecasting over time:} $\widehat{P}_{g^*}$ vs. fraction of trajectory length, across all trajectories. \Darkosp outperforms other methods and becomes more confident in the correct goal as the trajectories elapse.} \label{fig:auc_over_time}
\end{figure}

\subsection{Empirical Regret Analysis}

We empirically show that our model has no-regret with respect to the best model computed in hindsight under the MaxEntIRL loss function (negative log-loss). In particular, we compute the regret (cumulative loss difference) between our online algorithm and the best hindsight model using the batch MaxEntIOC algorithm \cite{ziebart2008maximum} at the end of all episodes. We plot the average regret $\frac{\mathcal{R}_t}{t}$ for each environment in the dataset in Figure~\ref{fig:empirical_regret}. The average regret of our algorithm approaches zero over time, matching our analysis. 

\begin{figure}[t]
\centering
 \includegraphics[height=1.1in]{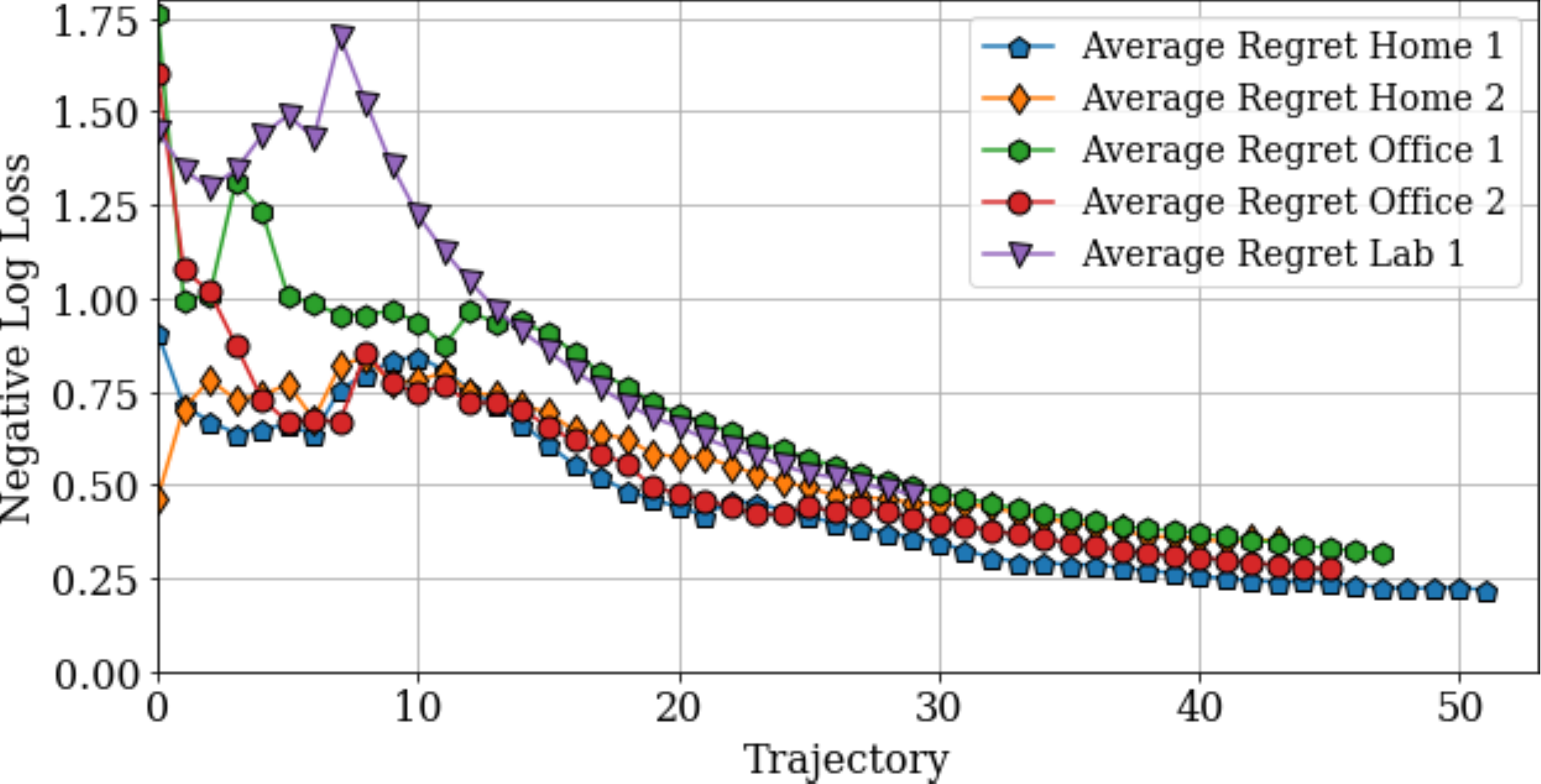}
\caption{\textbf{Empirical regret.} \Darkosp exhibits sublinear convergence in average regret. Initial noise is overcome after \Darkosp adjusts to the user's early behaviors.}\label{fig:empirical_regret}
\end{figure}

\subsection{Evaluation of Trajectory Length Estimates} \label{sec:exp_traj_len_forecasting}

Our model can also be used to estimate how long it will take a person to reach a predicted goal state. We predict the expected trajectory length as derived in Section~\ref{sec:future_state_and_subspace_prediction}. For the $n$-th episode, we use the normalized trajectory length prediction error defined as $\epsilon_n = \sum_{t=1}^{T_n} \frac{ | \tau_{nt} - \hat{\tau}_{nt} |}{ \tau_{nt} }$, where $\tau_{nt}$ is the true trajectory length and $\hat{\tau}_{nt}$ (Eq. \ref{eq:length}) is the predicted trajectory length. Proper evaluation of trajectory length towards a goal is challenging because our approach must learn valid goals in an online fashion. When a person approaches a new goal, our approach cannot accurately predict the goal because it has yet to learn that it is a valid goal state. As a result, our algorithm makes wrong goal predictions during episodes that terminate in new goal states. If we simply evaluate the mean performance, it will be dominated by the errors of the first episode terminating in a new goal state.

We evaluate median $\epsilon_n$ over all $N$ episodes. The median is not dominated by the errors of the first episode toward a new goal. We find most trajectory length forecasts are accurate, evidenced by the median of the normalized prediction error in Table~\ref{tab:exp_traj_stats}. We include a partial-trajectory nearest neighbors baseline (NN). In Lab 1, the median trajectory length estimate is within $6.3\%$ of the true trajectory length.

\begin{table}[t]
\centering
\footnotesize
\begin{tabular}{@{}llllll@{}}
\toprule
Statistic & Home 1 & Home 2 & Office 1 & Office 2 & Lab 1 \\
\midrule
Med. \% Err. & $30.0$ & $34.8$ & $17.3$ & $18.4$ & $6.3$ \\
Med. \% Err. NN & $29.0$ & $33.5$ & $42.9$ & $36.0$ & $35.4$ \\
\midrule
Mean $\left|\xi\right|$ & $20.5$ & $31.0$ & $27.1$ & $13.7$ & $23.5$\\
\bottomrule
\end{tabular}%
\caption{\textbf{Trajectory length forecasting results.} Error is relative to the true length of each trajectory. Most trajectory forecasts are fairly accurate.}\label{tab:exp_traj_stats} 
\end{table}

\section{Conclusion}

We proposed the first method for continuously modeling and forecasting a first-person camera wearer's future semantic behaviors at far-reaching spatial and temporal horizons. Our method goes beyond predicting the physical trajectory of the user to predict their future semantic goals, and models the user's relationship to objects and their environment. We have proposed several efficient and extensible methods for forecasting other semantic quantities of interest. Exciting avenues for future work include building upon the semantic state representation to model more aspects of the environment (which enables forecasting of more detailed futures), validation against human forecasting performance, and further generalizing the notion of a ``goal" and how goals are discovered.\\

\noindent{\textbf{Acknowledgements}:} We acknowledge the support of NSF NRI Grant 1227495 and JST CREST Grant JPMJCR14E1. The authors thank Drew Bagnell for technical discussion, Katherine Lagree and Gunnar A. Sigurdsson for data collection assistance, and the anonymous reviewers for their helpful comments.

\fi % if body

\clearpage

{\small
\bibliographystyle{ieee}
\bibliography{darko_bib}

\begin{thebibliography}{10}\itemsep=-1pt

\bibitem{abbeel2004apprenticeship}
P.~Abbeel and A.~Y. Ng.
\newblock Apprenticeship learning via inverse reinforcement learning.
\newblock In {\em Proceedings of the twenty-first international conference on
  Machine learning}, page~1. ACM, 2004.

\bibitem{Alahi_2016_CVPR}
A.~Alahi, K.~Goel, V.~Ramanathan, A.~Robicquet, L.~Fei-Fei, and S.~Savarese.
\newblock Social lstm: Human trajectory prediction in crowded spaces.
\newblock In {\em The IEEE Conference on Computer Vision and Pattern
  Recognition (CVPR)}, June 2016.

\bibitem{Cao_2013_CVPR}
Y.~Cao, D.~Barrett, A.~Barbu, S.~Narayanaswamy, H.~Yu, A.~Michaux, Y.~Lin,
  S.~Dickinson, J.~Mark~Siskind, and S.~Wang.
\newblock Recognize human activities from partially observed videos.
\newblock In {\em The IEEE Conference on Computer Vision and Pattern
  Recognition (CVPR)}, June 2013.

\bibitem{fathi2011understanding}
A.~Fathi, A.~Farhadi, and J.~M. Rehg.
\newblock Understanding egocentric activities.
\newblock In {\em 2011 International Conference on Computer Vision}, pages
  407--414. IEEE, 2011.

\bibitem{Furu:2010}
Y.~Furukawa, B.~Curless, S.~M. Seitz, and R.~Szeliski.
\newblock Towards internet-scale multi-view stereo.
\newblock In {\em CVPR}, 2010.

\bibitem{hoai2014max}
M.~Hoai and F.~De~la Torre.
\newblock Max-margin early event detectors.
\newblock {\em International Journal of Computer Vision}, 107(2):191--202,
  2014.

\bibitem{kitani2012activity}
K.~M. Kitani, B.~D. Ziebart, J.~A. Bagnell, and M.~Hebert.
\newblock Activity forecasting.
\newblock In {\em European Conference on Computer Vision}, pages 201--214.
  Springer, 2012.

\bibitem{koppula2016anticipating}
H.~S. Koppula and A.~Saxena.
\newblock Anticipating human activities using object affordances for reactive
  robotic response.
\newblock {\em IEEE transactions on pattern analysis and machine intelligence},
  38(1):14--29, 2016.

\bibitem{lan2014hierarchical}
T.~Lan, T.-C. Chen, and S.~Savarese.
\newblock A hierarchical representation for future action prediction.
\newblock In {\em European Conference on Computer Vision}, pages 689--704.
  Springer, 2014.

\bibitem{lee2012discovering}
Y.~J. Lee, J.~Ghosh, and K.~Grauman.
\newblock Discovering important people and objects for egocentric video
  summarization.
\newblock In {\em CVPR}, volume~2, page~7, 2012.

\bibitem{li2014prediction}
K.~Li and Y.~Fu.
\newblock Prediction of human activity by discovering temporal sequence
  patterns.
\newblock {\em IEEE transactions on pattern analysis and machine intelligence},
  36(8):1644--1657, 2014.

\bibitem{Li_2015_CVPR}
Y.~Li, Z.~Ye, and J.~M. Rehg.
\newblock Delving into egocentric actions.
\newblock In {\em The IEEE Conference on Computer Vision and Pattern
  Recognition (CVPR)}, June 2015.

\bibitem{ma2016going}
M.~Ma, H.~Fan, and K.~M. Kitani.
\newblock Going deeper into first-person activity recognition.
\newblock In {\em Proceedings of the IEEE Conference on Computer Vision and
  Pattern Recognition}, pages 1894--1903, 2016.

\bibitem{ma2016game}
W.-C. Ma, D.-A. Huang, N.~Lee, and K.~M. Kitani.
\newblock A game-theoretic approach to multi-pedestrian activity forecasting.
\newblock {\em arXiv preprint arXiv:1604.01431}, 2016.

\bibitem{mur2015orb}
R.~Mur-Artal, J.~Montiel, and J.~D. Tard{\'o}s.
\newblock Orb-slam: a versatile and accurate monocular slam system.
\newblock {\em IEEE Transactions on Robotics}, 31(5):1147--1163, 2015.

\bibitem{pirsiavash2012detecting}
H.~Pirsiavash and D.~Ramanan.
\newblock Detecting activities of daily living in first-person camera views.
\newblock In {\em Computer Vision and Pattern Recognition (CVPR), 2012 IEEE
  Conference on}, pages 2847--2854. IEEE, 2012.

\bibitem{platt1999probabilistic}
J.~Platt et~al.
\newblock Probabilistic outputs for support vector machines and comparisons to
  regularized likelihood methods.
\newblock {\em Advances in large margin classifiers}, 10(3):61--74, 1999.

\bibitem{ratliff2006maximum}
N.~D. Ratliff, J.~A. Bagnell, and M.~A. Zinkevich.
\newblock Maximum margin planning.
\newblock In {\em Proceedings of the 23rd international conference on Machine
  learning}, pages 729--736. ACM, 2006.

\bibitem{redmon2016you}
J.~Redmon, S.~Divvala, R.~Girshick, and A.~Farhadi.
\newblock You only look once: Unified, real-time object detection.
\newblock In {\em Proceedings of the IEEE Conference on Computer Vision and
  Pattern Recognition}, pages 779--788, 2016.

\bibitem{rhinehart2016learning}
N.~Rhinehart and K.~M. Kitani.
\newblock Learning action maps of large environments via first-person vision.
\newblock In {\em Proceedings of the IEEE Conference on Computer Vision and
  Pattern Recognition}, pages 580--588, 2016.

\bibitem{ryoo2011human}
M.~S. Ryoo.
\newblock Human activity prediction: Early recognition of ongoing activities
  from streaming videos.
\newblock In {\em Computer Vision (ICCV), 2011 IEEE International Conference
  on}, 2011.

\bibitem{ryoo2013first}
M.~S. Ryoo and L.~Matthies.
\newblock First-person activity recognition: What are they doing to me?
\newblock In {\em Proceedings of the IEEE Conference on Computer Vision and
  Pattern Recognition}, pages 2730--2737, 2013.

\bibitem{shalev2012online}
S.~Shalev-Shwartz et~al.
\newblock Online learning and online convex optimization.
\newblock {\em Foundations and Trends{\textregistered} in Machine Learning},
  4(2):107--194, 2012.

\bibitem{simonyan2014very}
K.~Simonyan and A.~Zisserman.
\newblock Very deep convolutional networks for large-scale image recognition.
\newblock {\em arXiv preprint arXiv:1409.1556}, 2014.

\bibitem{Park_2016_CVPR}
H.~Soo~Park, J.-J. Hwang, Y.~Niu, and J.~Shi.
\newblock Egocentric future localization.
\newblock In {\em The IEEE Conference on Computer Vision and Pattern
  Recognition (CVPR)}, June 2016.

\bibitem{su2016social}
S.~Su, J.~P. Hong, J.~Shi, and H.~S. Park.
\newblock Social behavior prediction from first person videos.
\newblock {\em arXiv preprint arXiv:1611.09464}, 2016.

\bibitem{sun2016learning}
W.~Sun, A.~Venkatraman, B.~Boots, and J.~A. Bagnell.
\newblock Learning to filter with predictive state inference machines.
\newblock In {\em Proceedings of The 33rd International Conference on Machine
  Learning}, pages 1197--1205, 2016.

\bibitem{Vondrick_2016_CVPR}
C.~Vondrick, H.~Pirsiavash, and A.~Torralba.
\newblock Anticipating visual representations from unlabeled video.
\newblock In {\em The IEEE Conference on Computer Vision and Pattern
  Recognition (CVPR)}, June 2016.

\bibitem{walker2014patch}
J.~Walker, A.~Gupta, and M.~Hebert.
\newblock Patch to the future: Unsupervised visual prediction.
\newblock In {\em 2014 IEEE Conference on Computer Vision and Pattern
  Recognition}, pages 3302--3309. IEEE, 2014.

\bibitem{wei1994time}
W.~W.-S. Wei.
\newblock {\em Time series analysis}.
\newblock Addison-Wesley publ Reading, 1994.

\bibitem{xie2013inferring}
D.~Xie, S.~Todorovic, and S.-C. Zhu.
\newblock Inferring ``dark matter" and ``dark energy" from videos.
\newblock In {\em Proceedings of the IEEE International Conference on Computer
  Vision}, pages 2224--2231, 2013.

\bibitem{zhou2014learning}
B.~Zhou, A.~Lapedriza, J.~Xiao, A.~Torralba, and A.~Oliva.
\newblock Learning deep features for scene recognition using places database.
\newblock In {\em Advances in neural information processing systems}, pages
  487--495, 2014.

\bibitem{ziebart2010modeling}
B.~D. Ziebart.
\newblock {\em Modeling Purposeful Adaptive Behavior with the Principle of
  Maximum Causal Entropy}.
\newblock PhD thesis, Carnegie Mellon University, 2010.

\bibitem{ziebart2008maximum}
B.~D. Ziebart, A.~L. Maas, J.~A. Bagnell, and A.~K. Dey.
\newblock Maximum entropy inverse reinforcement learning.
\newblock In {\em AAAI Conference on Artificial Intelligence}, pages
  1433--1438, 2008.

\bibitem{ziebart2009planning}
B.~D. Ziebart, N.~Ratliff, G.~Gallagher, C.~Mertz, K.~Peterson, J.~A. Bagnell,
  M.~Hebert, A.~K. Dey, and S.~Srinivasa.
\newblock Planning-based prediction for pedestrians.
\newblock In {\em 2009 IEEE/RSJ International Conference on Intelligent Robots
  and Systems}, pages 3931--3936. IEEE, 2009.

\end{thebibliography}
}

\ifbuildappendix
\clearpage
\onecolumn
\appendix
\section*{Appendix}

\section{Reward Function Feature Ablation Analysis}

\begin{table}[tb]
\centering
\footnotesize
\begin{tabular}{@{}lllllll@{}}
\toprule
Feature Type & Home 1 & Home 2 & Office 1 & Office 2 & Lab 1 \\
\midrule
State+Action & $\mathbf{0.851}$ & $\mathbf{0.683}$  & $\mathbf{0.700}$ & $\mathbf{0.666}$ & $0.880$ \\
State only &$ 0.735$ & $0.574$ & $0.581$ & $0.549$ & $\mathbf{0.892}$ \\
Position only &$ 0.674$ & $0.597$ & $0.605$ & $0.622$ & $0.886$ \\
\bottomrule
\end{tabular}
\caption{\textbf{Feature Ablation Results:} Full state and action features (Sec. \ref{sec:terminology}) yield best goal prediction results.} \label{tab:ablation} 
\end{table}

In Table~\ref{tab:ablation}, we show the mean true goal probability when labels are used as detectors (to isolate performance in the ideal case). While the purely positional representation of state performs well, it is almost always outperformed by the full representation of rewards that include features of both the full state and action. In Lab 1, the simpler representations slightly outperform the full, due to the relative simplicity of the high-level activities in Lab 1. Here, knowledge of the state and previous goal alone is highly predictive of future goal. 

\section{Incorporating Detection Noise} \label{sec:goal_det_forecasting}

Current paradigms in vision often yield noise in the action and goal detectors necessary for \Darko. We first describe our method for incorporating uncertainty in each goal detection, then conduct a performance analysis with synthetic noise. Then, we analyze the performance with real, noisy goal detection. We find \Darkosp can still perform well with forms of noisy goal and action detection. We find incorporating goal uncertainty significantly improves performance with synthetic noise, and shows improvements in the real goal detector setting. These results show that \Darkosp can tolerate the effects of noise, and further support the claim that it can enjoy the benefits of better scene and activity detection algorithms.

\noindent\textbf{Harnessing goal detection confidence:} In many scenarios, probability $\rho_g \in [0, 1]$ may be associated with each goal detection. We designed an effective method for handling real-world uncertainty. For known perfect goals, \textsc{SoftValueIteration} uses $V(g) = 0, \forall g \in \mathcal{S}_g$. Each goal is a maxima of $V(s) \in (-\infty, 0], \forall s \in \mathcal{S}$ and represents a reward of $1$ in log space. \textit{We replace each goal value with its log-probability: $V(g) = \ln{\rho_g}$,} \textit{which has the effect of biasing the policy towards goals with greater certainty}. This results from the value iteration assigning higher value to states and actions closer to more certain goals, which makes the policy likelier to visit them. For example, if the goal detector yields a false positive of \texttt{bathroom} in the same area as a true positive detection of \texttt{kitchen}, the goal prediction posteriors for both goals will suffer, unless the false positive has an associated low $\rho_g$ (high uncertainty), in which case the policy is biased towards the correct goal of kitchen.  

\begin{figure}[t]
\centering
\includegraphics[height=1.4in]{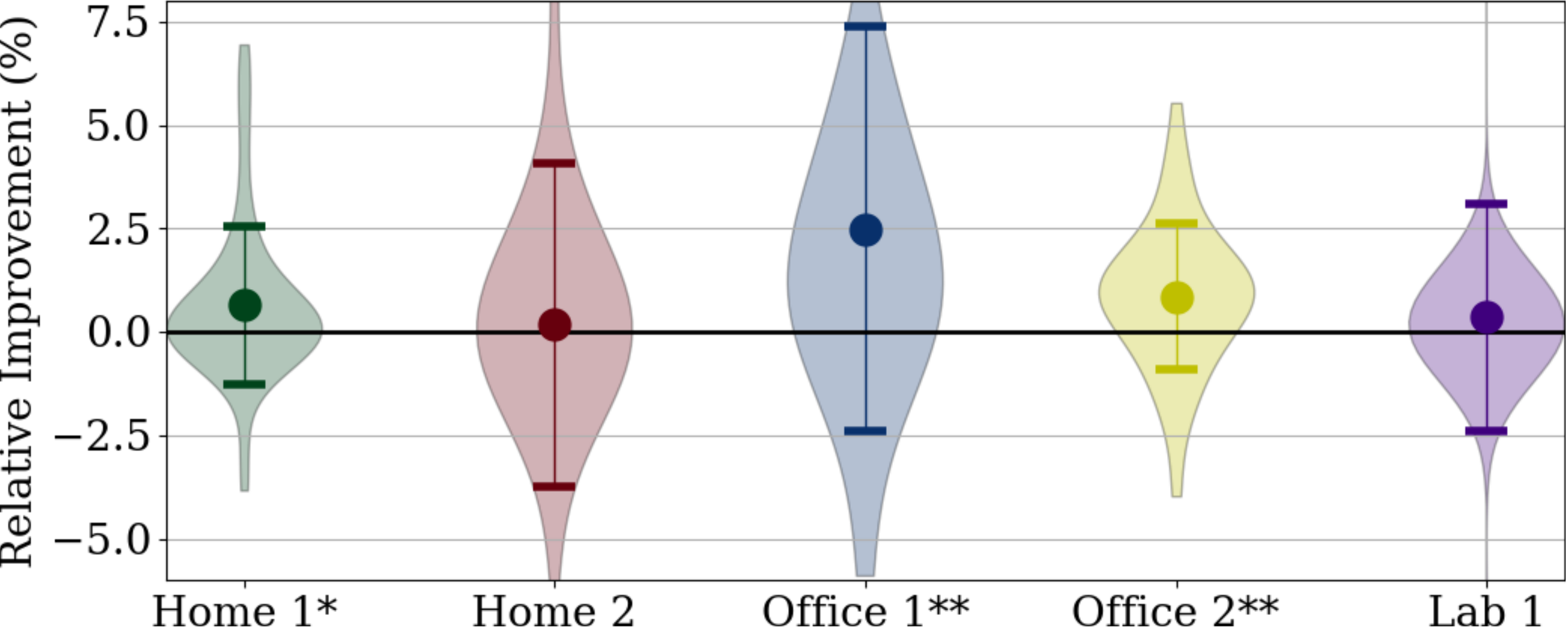}
\caption{\textbf{Relative improvement from incorporating goal uncertainty.} Per-scene violin plots, means, and standard deviations are shown. Per-scene one-sided paired t-tests are performed, testing the hypotheses that incorporating goal uncertainty improves goal prediction performance. A \textsuperscript{*} indicates $p < 0.05$, and \textsuperscript{**} indicates $p < 0.005$.}\label{fig:synth_per_scene}
\end{figure}

\noindent\textbf{Noise analysis:} We first analyze \Darkosp under the effect of adding noise to the GT. We add incorrect goal detections with probabilities $\rho_g\sim\mathcal{N}(0.1,0.05)$, under various amounts of noise inserted uniformly at random across time: from $10\%, 20\%, \dots, 90\%$ of the number of original goal detections. For every scene, at each noise amount, we sample noise 5 times, and run \Darkosp with and without goal uncertainty for each corrupted sample, resulting in 225 paired experiments that evaluate the average goal forecasting probability. Per-scene results are shown in Figures~\ref{fig:synth_per_scene}. \textit{A one-sided Wilcoxon signed-rank test supports the hypothesis that incorporating high goal uncertainty yields better goal posterior prediction performance than not incorporating the uncertainty with $p<10^{-7}$}.

\section{Proof of Regret Bound} \label{sec:regret_proof}
Our regret bound is:
\begin{align}
    \mathcal{R}_t \leq 2B\sqrt{2td}, \label{eq:regret}
\end{align}
where $B$ is a bound on the norm of $\theta$, $d$ is feature dimensionality, and $t$ is the episode count (regret after the $t$'th episode).

\begin{proof} By Equation 2.5 of \cite{shalev2012online}, the regret of
online gradient descent is bounded:

\begin{align}
    \mathcal{R}_t \leq \frac{1}{2\lambda}\|\theta\|_2^2 + \lambda \sum_{i=1}^t\|\nabla_{\theta_t}\|_2^2. \label{eq:regret_shalev}
\end{align}

By using bounds on $\|\theta\|_2^2, \|\nabla_{\theta_t}\|_2^2$, and a minimizing choice of $\lambda$, we will prove the result.

Writing the general gradient in terms of the expected features (and omitting the subscript $t$):
\begin{align}
    \|\nabla_{\theta}\|_2^2 &= \|\overline{f}-\hat{f}\|_2^2 \nonumber \\
    &= \fbar^T\fbar + \fhat^T\fhat - 2\fbar^T\fhat \label{eq:nabla_pt1}
\end{align}

Using: 
\begin{align}
0&\leq \|x-y\|_2^2 = x^Tx + y^Ty - 2x^Ty  \nonumber \\
2x^Ty &\leq x^Tx + y^Ty \nonumber \\
2(-x)^Ty &\leq (-x)^T(-x) + y^Ty \nonumber \\
-2x^Ty &\leq x^Tx + y^Ty, \nonumber \\
\therefore -2\fbar^T\fhat &\leq \fbar^T\fbar + \fhat^T\fhat,~(\text{Setting}~ x=\fbar, y=\fhat) \nonumber
\end{align}

then Equation~\ref{eq:nabla_pt1} becomes:
\begin{align}
 \|\nabla_{\theta}\|_2^2 &\leq \fbar^T\fbar + \fhat^T\fhat + \fbar^T\fbar + \fhat^T\fhat  \nonumber\\
 &= 2\fbar^T\fbar + 2\fhat^T\fhat \nonumber\\
 &\leq 4d.~(\text{Since}~\fbar,\fhat\in[0,1]^d)\label{eq:theta_bound}
\end{align}

Thus, using Equation~\ref{eq:theta_bound} in Equation~\ref{eq:regret_shalev}, and that the projection step of $\theta$ (constraining the set of $\theta$ to be the convex ball with radius $B$) ensures $\|\theta\|_2 \leq B$:
\begin{align}
    \mathcal{R}_t &\leq \frac{B^2}{2\lambda} + \lambda \sum_{i=1}^t4d \nonumber\\
   &=  \frac{B^2}{2\lambda} + 4\lambda td. \nonumber
\end{align}

With the minimizing choice of $\lambda = \frac{B}{2\sqrt{2td}}$,
\begin{align}
    \mathcal{R}_t &\leq B\sqrt{2td} + \frac{2Btd}{\sqrt{2td}}= 2B\sqrt{2td} \nonumber
\end{align}
\end{proof}

\section{Derivation of Other Inference Tasks} \label{sec:action_subpsace}

\subsection{Action-Subspace Visitation}
To derive the action-subspace visitation, we first use the posterior expected visitation count of performing an action $a_y$ immediately after arriving at a state $s_x$ is given in Equation~\ref{eq:act-state_definition}, from \cite{ziebart2010modeling}.

\begin{align}
D_{a_y,s_x | \xi_{0 \rightarrow t}} &\triangleq  \mathbb{E}_{P(\xi_{t+1 \rightarrow T} | \xi_{0 \rightarrow t})}\left[\sum_{\tau = t+1}^TI(s_{\tau} = s_x)I(a_\tau = a_y) \right] \label{eq:act-state_definition}\\
D_{a_y,s_x | \xi_{0 \rightarrow t}} &= \pi(a_y | s_x)D_{s_x | \xi_{0 \rightarrow t}} \label{eq:d_a_policy}
\end{align}

Our definition of the posterior expected action subspace visitation count is given in Equation~\ref{eq:exp_a_subspace}. This expresses the future expectation the
user will perform an action $a_y$ while in a subspace $S_p$, given their current trajectory $\xi_{0\rightarrow t}$.
\begin{align}
D_{a_y,\mathcal{S}_p | \xi_{0 \rightarrow t}} &\triangleq \mathbb{E}_{P(\xi_{t+1 \rightarrow T} | \xi_{0 \rightarrow t})}\left[\sum_{\tau = t+1}^T I(s_{\tau} \in \mathcal{S}_p)I(a_\tau = a_y) \right] \label{eq:exp_a_subspace}\\
&= \mathbb{E}_{P(\xi_{t+1 \rightarrow T} | \xi_{0 \rightarrow t})}\left[\sum_{s_x \in \mathcal{S}_p}\sum_{\tau = t+1}^T I(s_{\tau} = s_x)I(a_\tau = a_y) \right] \nonumber \\
&= \sum_{s_x \in \mathcal{S}_p}\mathbb{E}_{P(\xi_{t+1 \rightarrow T} | \xi_{0 \rightarrow t})}\left[\sum_{\tau = t+1}^T I(s_{\tau} = s_x)I(a_\tau = a_y) \right] \nonumber \\
&= \sum_{s_x \in \mathcal{S}_p} D_{a_y,s_x | \xi_{0 \rightarrow t}} \nonumber \\
&= \sum_{s_x \in \mathcal{S}_p} \pi(a_y | s_x) D_{s_x | \xi_{0 \rightarrow t}} 
\end{align}.

Thus, the posterior expected action subspace visitation is straightforward to compute with $D_{s_x | \xi_{0\rightarrow t}}$. Various inference tasks can be constructed by choosing $a_y$ and $\mathcal{S}_p$ appropriately.

\subsection{Joint Action-State Subspace Visitation}
We additionally derive the expected transition count from a subspace of states to a subspace of actions. This expresses the expectation that the user will perform an $a_y \in \mathcal{A}_y$ from a $s_x \in \mathcal{S}_p$. It is defined as:

\begin{align}
D_{\mathcal{A}_y,\mathcal{S}_p | \xi_{0 \rightarrow t}} &\triangleq \mathbb{E}_{P(\xi_{t+1 \rightarrow T} | \xi_{0 \rightarrow t})}\left[\sum_{\tau = t+1}^T I(s_{\tau} \in \mathcal{S}_p)I(a_\tau \in \mathcal{A}_y) \right] \label{eq:exp_joint_subspace}\\
&= \mathbb{E}_{P(\xi_{t+1 \rightarrow T} | \xi_{0 \rightarrow t})}\left[\sum_{a_y \in \mathcal{A}_y}\sum_{s_x \in \mathcal{S}_p}\sum_{\tau = t+1}^T I(s_{\tau} = s_x)I(a_\tau = a_y) \right] \nonumber \\
&= \sum_{a_y \in \mathcal{A}_y}\sum_{s_x \in \mathcal{S}_p}\mathbb{E}_{P(\xi_{t+1 \rightarrow T} | \xi_{0 \rightarrow t})}\left[\sum_{\tau = t+1}^T I(s_{\tau} = s_x)I(a_\tau = a_y) \right] \nonumber \\
&= \sum_{a_y \in \mathcal{A}_y} \sum_{s_x \in \mathcal{S}_p}D_{a_y,s_x | \xi_{0 \rightarrow t}} \nonumber \\
&= \sum_{a_y \in \mathcal{A}_y}\sum_{s_x \in \mathcal{S}_p} \pi(a_y | s_x) D_{s_x | \xi_{0 \rightarrow t}}.
\end{align}

Again, computing this quantity is straightforward with $D_{s_x | \xi_{0\rightarrow t}}$. By marginalizing $D_{s_x |\xi_{0 \rightarrow}}$ over various action and state subspaces that have semantic meaning, different inference quantities can be expressed and computed.

\section{Additional Dataset Information}
\noindent{\textbf{Objects: }} The set of objects available in each environment is shown in Table~\ref{tab:environment_objects}. \\
\noindent{\textbf{Scene types: }} The set of scene types in each environment is shown in Table~\ref{tab:environment_scenes}. \\
\noindent{\textbf{Labels: }} A small snippet of ground truth for Home 1 is shown in Table~\ref{tab:home1_gt_snippet}. The ground truth pairs frame indices (timestamps) with actions and goal arrivals.

\begin{table}[t]
\centering
\begin{tabular}{@{}ll@{}}
\toprule
 Environment & Object Set \\
 \midrule
Home 1 & $\{\text{bookbag, book, blanket, coat, laptop, mug, plate, snack, towel}\}$ \\ 
Home 2 & $\{\text{bookbag, book, blanket, coat, guitar, laptop, mug, plate, remote, snack, towel}\}$ \\
Office 1 & $\{\text{bookbag, textbook, bottle, coat, laptop, mug, paper, plate, snack}\}$ \\
Office 2 & $\{\text{bookbag, textbook, bottle, coat, laptop, mug, paper, plate, snack}\}$ \\
Lab 1 & $\{\text{beaker, coat, plate, pipette, tube}\}$ \\
\bottomrule
\end{tabular} 
\caption{\textbf{Objects available in each environment.}} \label{tab:environment_objects}
\end{table}

\begin{table}[t]
\centering
\begin{tabular}{@{}ll@{}}
\toprule

Environment & Scene Type Set \\
 \midrule
Home 1 & $\{\text{bathroom, bedroom, exit, dining room, kitchen, living room,  office}\}$ \\ 
Home 2 & $\{\text{bathroom, bedroom, exit, dining room, kitchen, living room,  office, television stand}\}$ \\
Office 1 & $\{\text{bathroom, exit, kitchen, lounge, office, printer station, water fountain}\}$ \\

Office 2 & $\{\text{bathroom, exit, kitchen, lounge, office, printer room, water fountain}\}$\\

Lab 1 & $\{\text{cabinet stand, exit, gel room, lab bench 1, lab bench 2, refrigeration room}\}$ \\
\bottomrule
\end{tabular} 

\caption{\textbf{Scene types available in each environment.}} \label{tab:environment_scenes}
\end{table}

\begin{table}[t]
\centering
\begin{tabular}{@{}l|llllll@{}}
\toprule
Frame Index & 6750 & 6900 & 7200 & 7400 & 7630 &7700\\
 Action/Arrival  & Release Coat  & Acquire Bookbag &  Arrive Office & Acquire Mug & Arrive Kitchen & Release Mug\\
 \bottomrule
\end{tabular} 
\caption{\textbf{Labels example:} A small snippet of ground truth labels for Home 1.} \label{tab:home1_gt_snippet}
\end{table}

\section{Regret with Detectors}
We additionally show the empirical regret when using our goal discovery and action detection methods in Figure~\ref{fig:empirical_regret_fa}. We observe somewhat noisier regret behavior than in the original case, as the underlying demonstrations are noisier. The number of trajectories in Office 2 is higher here due to errors in the goal forecasting method, resulting in more goals being detected, which segments the demonstrations into more trajectories.

\begin{figure}[t]
\centering
\includegraphics[height=1.5in]{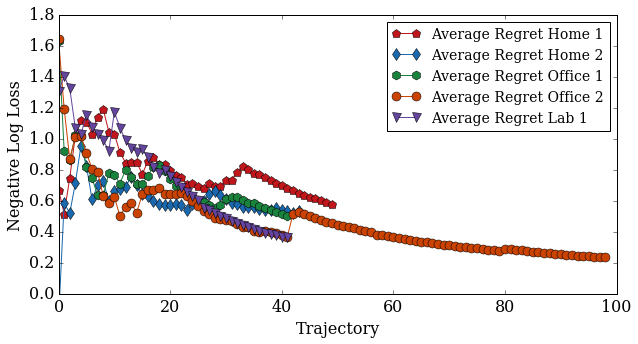}
\caption{\textbf{Noisy Empirical Regret.} \Darko's online behavior model exhibits sublinear convergence in average regret. Initial noise is overcome after \Darkosp adjusts to learning about the user's early behaviors.}\label{fig:empirical_regret_fa}
\end{figure}

\section{Visualizations}
We provide example 3D visualizations of 1) goal posterior 2) future state visitation and 3) the value function.

\subsection{Goal posterior visualization}
To emphasize our empirical finding that modeling the interaction with objects is useful for goal posterior prediction, we show an example sequence of predictions in Figure~\ref{fig:mug_overhead_posterior}. Before the user grabs the mug, our algorithm predicts roughly equivalent probability to both the bedroom and kitchen. We see that after the user grabs the mug, \Darkosp has high confidence that the user will go to the kitchen.
\begin{figure*}[t]
\begin{subfigure}[c]{.45\textwidth}
\centering
\includegraphics[trim={0cm 2.25cm 0cm 2.25cm}, clip,height=2in]{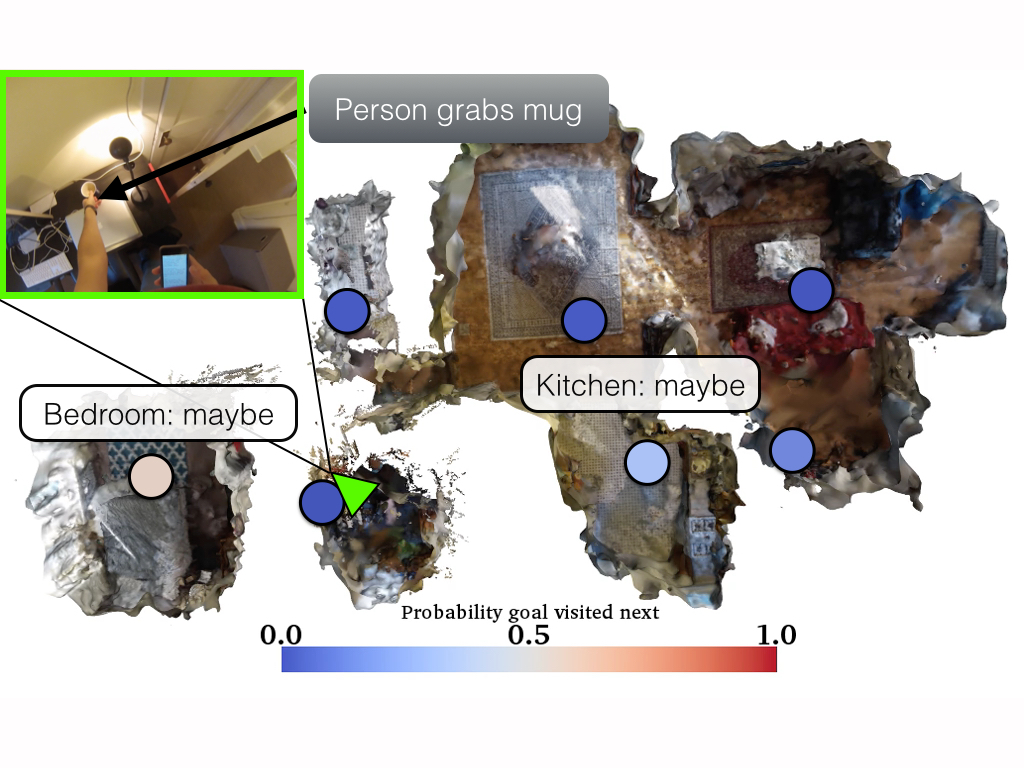}
\caption{Goal Posterior Before Mug Acquired} \label{subfig:mug_posterior_1}
\end{subfigure}
\begin{subfigure}[c]{.55\textwidth}
\centering
\includegraphics[trim={0cm 2.25cm 0cm 2.25cm}, clip,height=2in]{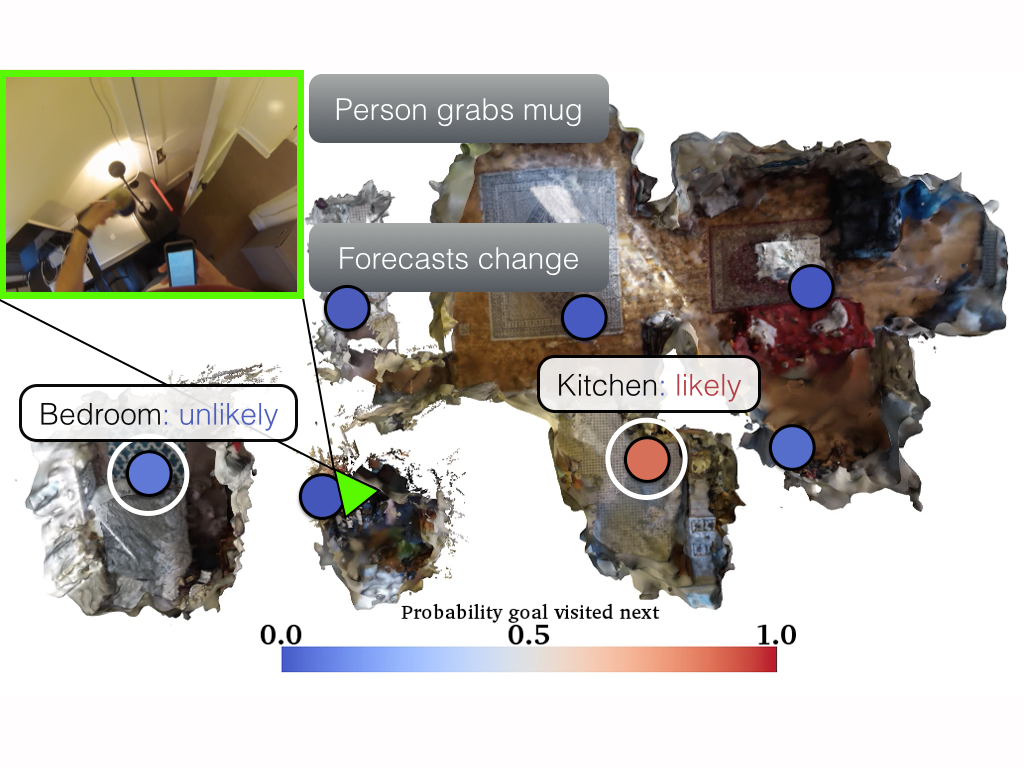}
\caption{Goal Posterior After Mug Acquired} \label{subfig:mug_posterior_2}
\end{subfigure}
\vspace{-1mm}
\caption{\textbf{Goal Posterior Change Visualization:} Goal posteriors for two frames are visualized in the Home 1 environment. The person's location is in green, images from the camera are inset at top left, and goal posteriors are colored according to the above colormaps. Before grabbing the mug (Figure~\ref{subfig:mug_posterior_1}), \Darkosp forecasts roughly equivalent probability to bedroom and kitchen. After the user grabs the mug (Figure~\ref{subfig:mug_posterior_2}), \Darkosp correctly predicts the user is likeliest to go to the kitchen.} \label{fig:mug_overhead_posterior}\vspace{-3mm}
\end{figure*}

\subsection{Future state visitation visualizations} \label{sec:fsv}
See Figure~\ref{fig:fsv} for example visualizations of the expected future visitation counts. In order to visualize in 3 dimensions, we take the max visitation count across all states at each position. In rows 1, 3, and 4, a single demonstration is shown, which adapts to the agent's trajectory (history). In row 2, the future state distribution drastically changes after each time the agent reaches a new goal.

\begin{figure}[ht]
\centering
\begin{subfigure}[c]{.33\textwidth}
\centering
\includegraphics[width=\textwidth]{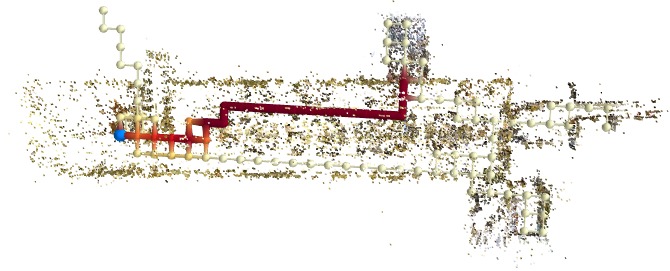}
\end{subfigure}
\begin{subfigure}[c]{.33\textwidth}
\centering
\includegraphics[width=\textwidth]{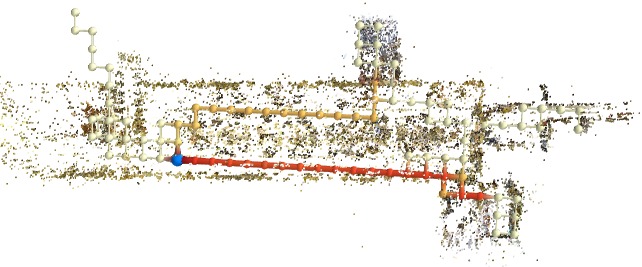}
\end{subfigure}
\begin{subfigure}[c]{.33\textwidth}
\centering
\includegraphics[width=\textwidth]{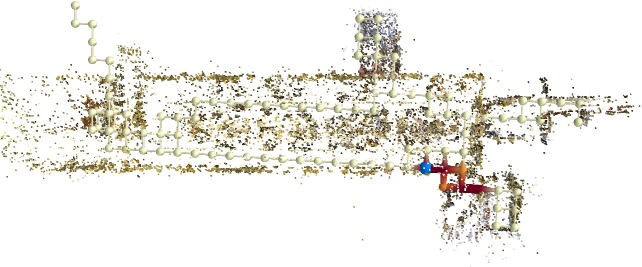}
\end{subfigure}
\begin{subfigure}[c]{.33\textwidth}
\centering
\includegraphics[width=\textwidth]{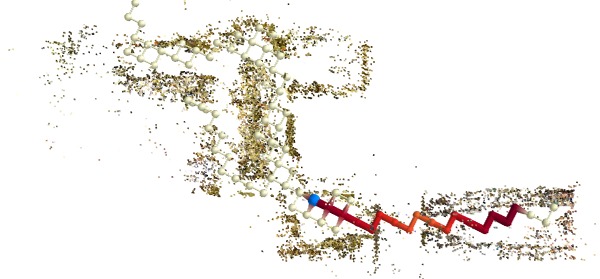}
\end{subfigure}
\begin{subfigure}[c]{.33\textwidth}
\centering
\includegraphics[width=\textwidth]{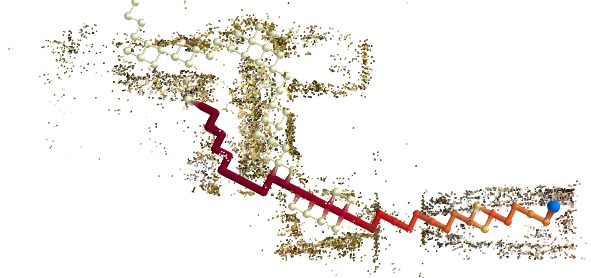}
\end{subfigure}
\begin{subfigure}[c]{.33\textwidth}
\centering
\includegraphics[width=\textwidth]{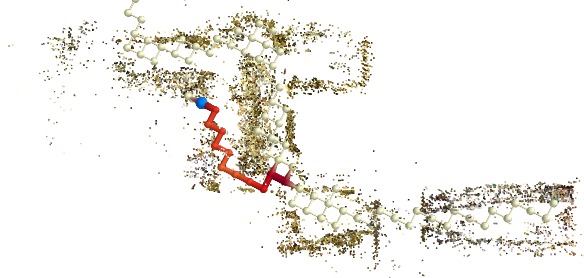}
\end{subfigure} %%
\begin{subfigure}[c]{.33\textwidth}
\centering
\includegraphics[width=\textwidth]{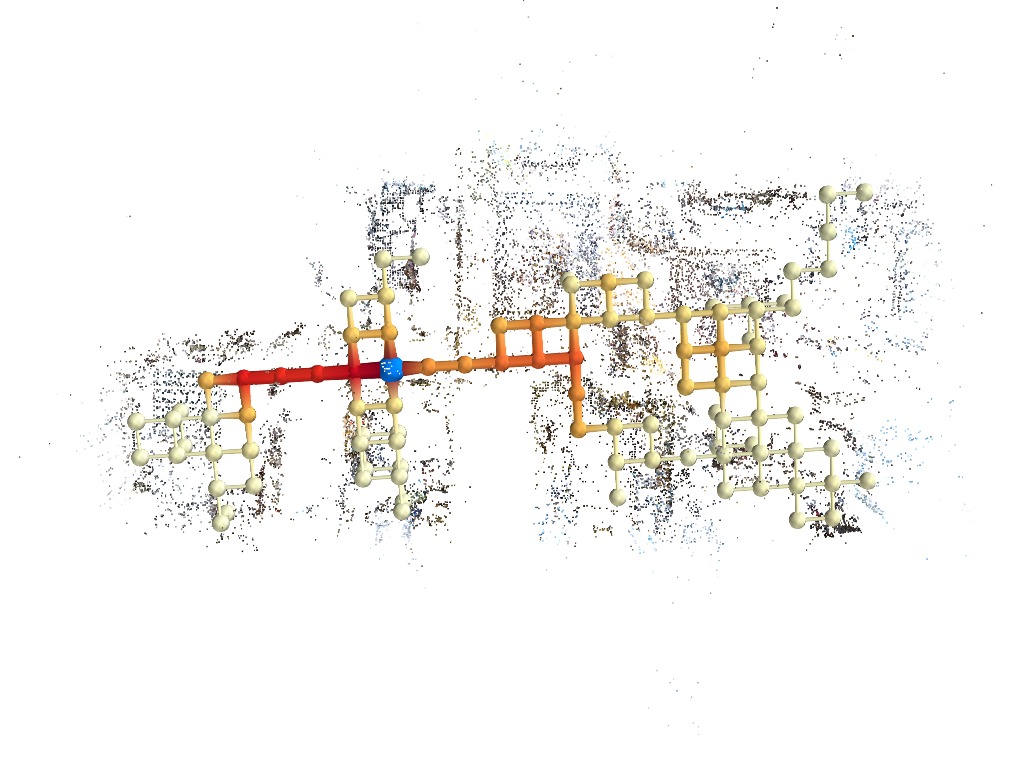}
\end{subfigure}
\begin{subfigure}[c]{.33\textwidth}
\centering
\includegraphics[width=\textwidth]{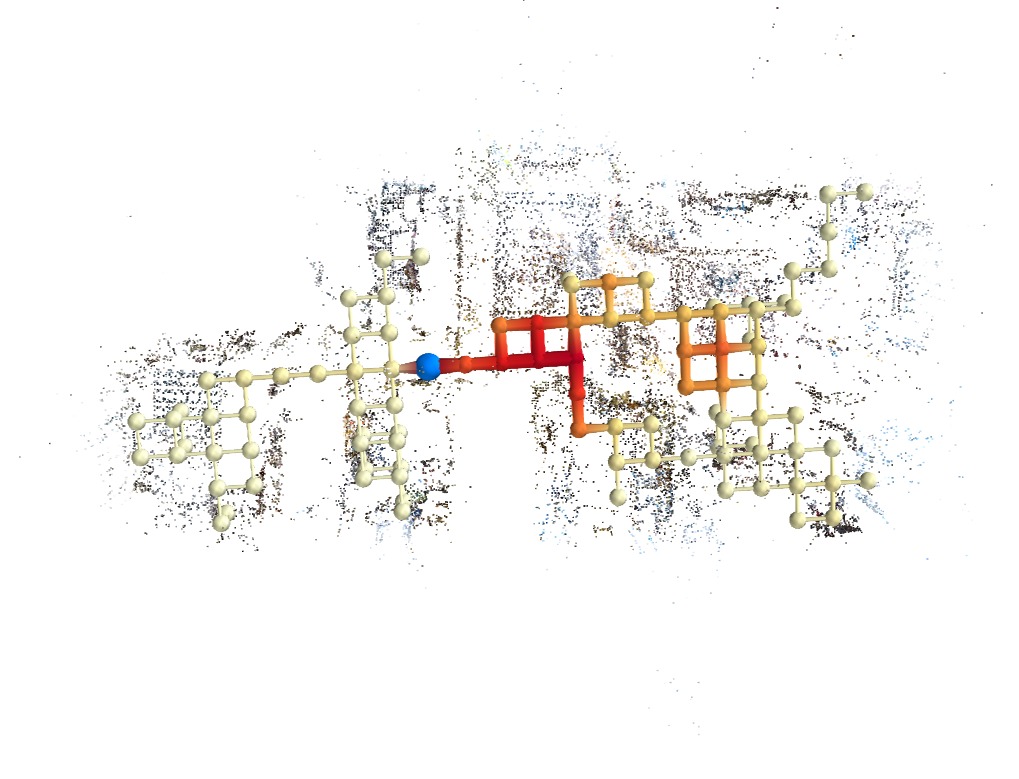}
\end{subfigure}
\begin{subfigure}[c]{.33\textwidth}
\centering
\includegraphics[width=\textwidth]{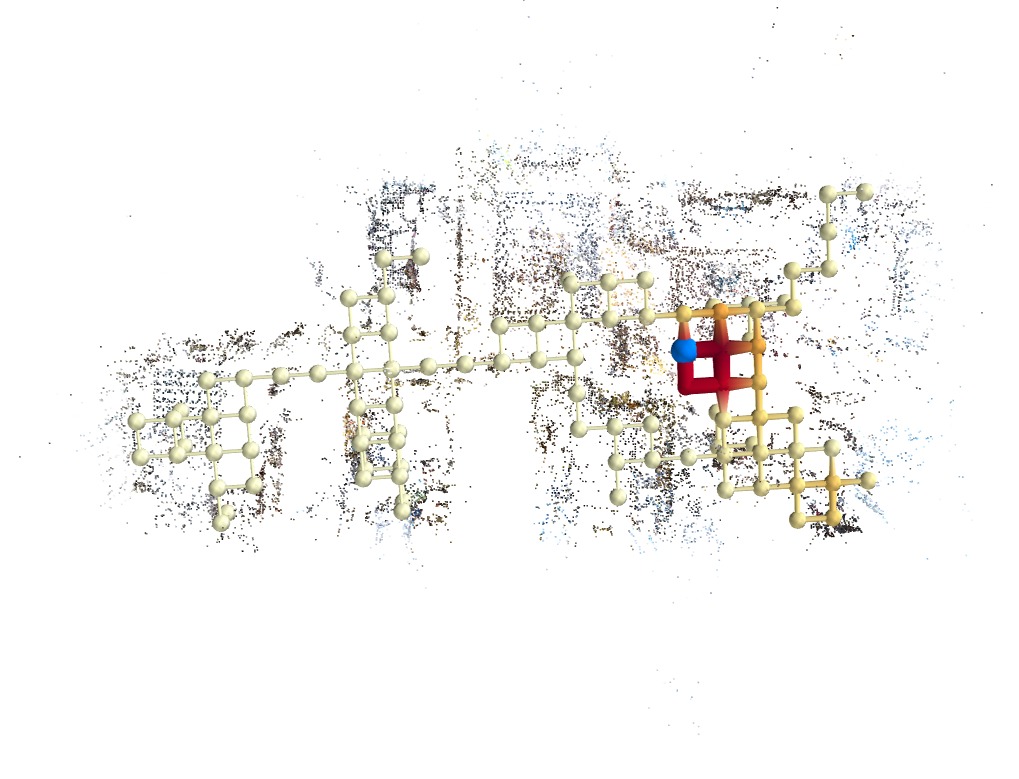}
\end{subfigure} %%
\begin{subfigure}[c]{.30\textwidth}
\centering
\includegraphics[width=\textwidth]{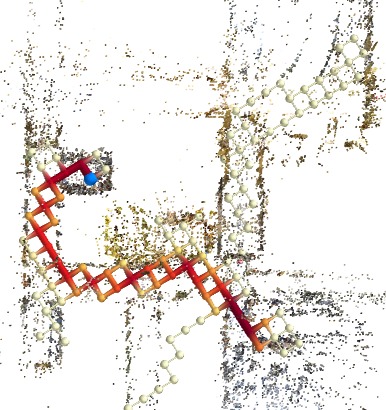}
\end{subfigure}
\begin{subfigure}[c]{.30\textwidth}
\centering
\includegraphics[width=\textwidth]{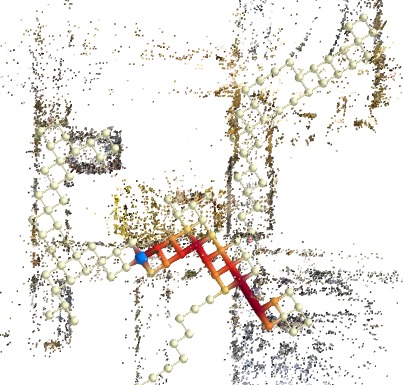}
\end{subfigure}
\begin{subfigure}[c]{.30\textwidth}
\centering
\includegraphics[width=\textwidth]{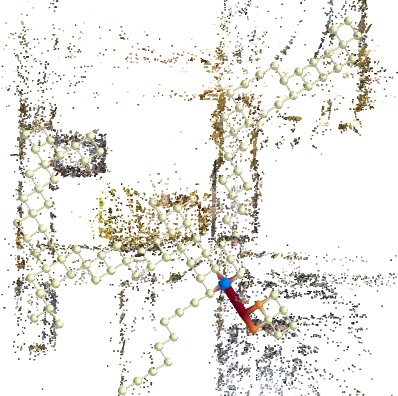}
\end{subfigure}
\caption{Future state visitation predictions changing as the agent (blue sphere) follows their trajectory. The state visitations are projected to 3D by taking the max over all states at each location. The visualizations are, by row: Office 1, Lab 1, Home 1., Office 2} \label{fig:fsv}
\end{figure}

\subsection{Value function visualizations} \label{sec:value}
See Figure~\ref{fig:value} for example visualizations of the value function over time. Note 1) the state space size changes, and 2) that the value function changes over time, as the component of state that indicates the previous goal affects the value function.

\begin{figure}[ht]
\centering
\begin{subfigure}[c]{.33\textwidth}
\centering
\includegraphics[width=\textwidth]{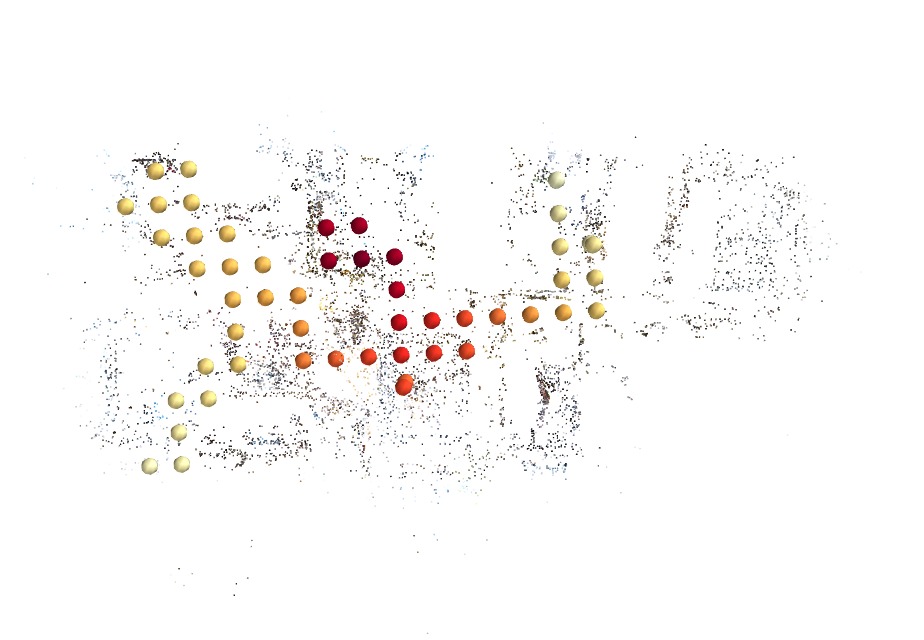}
\end{subfigure}
\begin{subfigure}[c]{.33\textwidth}
\centering
\includegraphics[width=\textwidth]{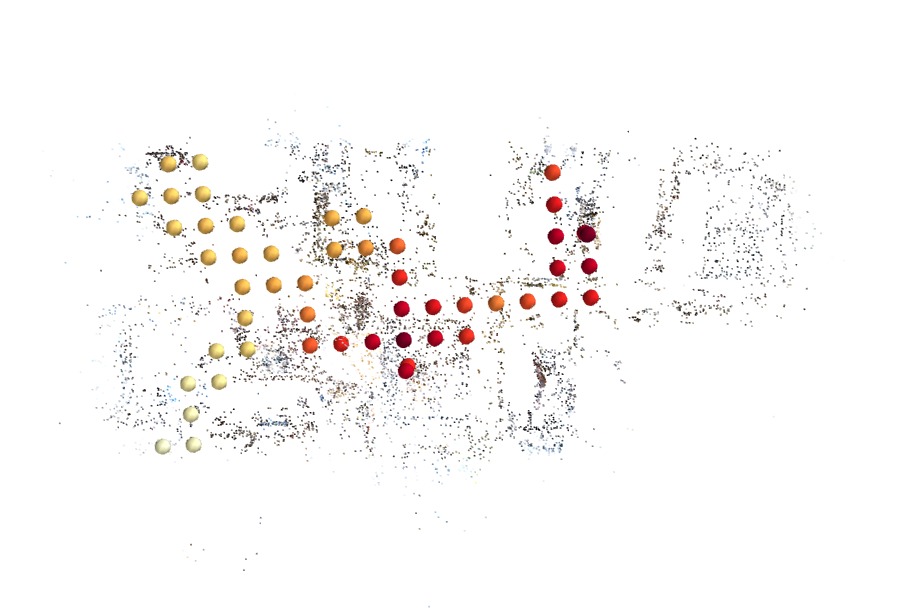}
\end{subfigure}
\begin{subfigure}[c]{.33\textwidth}
\centering
\includegraphics[width=\textwidth]{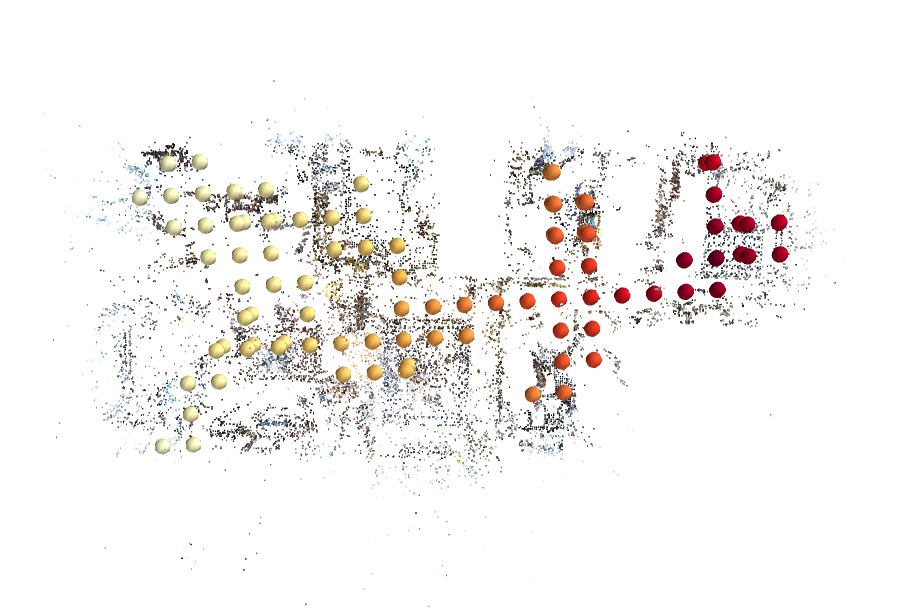}
\end{subfigure}
\begin{subfigure}[c]{.33\textwidth}
\centering
\includegraphics[width=\textwidth]{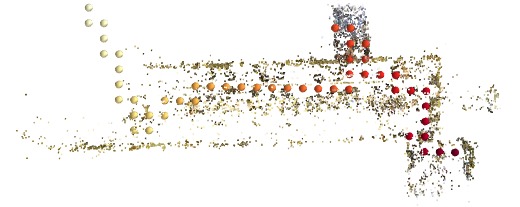}
\end{subfigure}
\begin{subfigure}[c]{.33\textwidth}
\centering
\includegraphics[width=\textwidth]{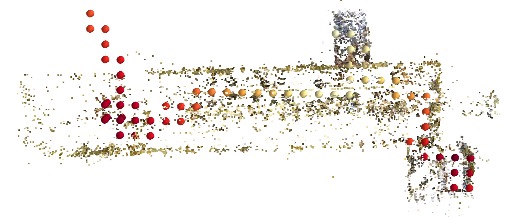}
\end{subfigure}
\begin{subfigure}[c]{.33\textwidth}
\centering
\includegraphics[width=\textwidth]{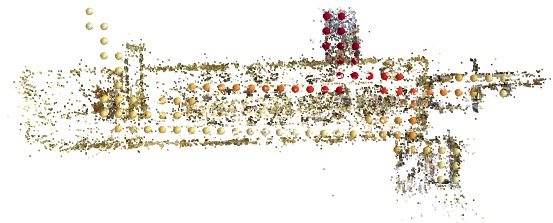}
\end{subfigure}
\begin{subfigure}[c]{.33\textwidth}
\centering
\includegraphics[width=\textwidth]{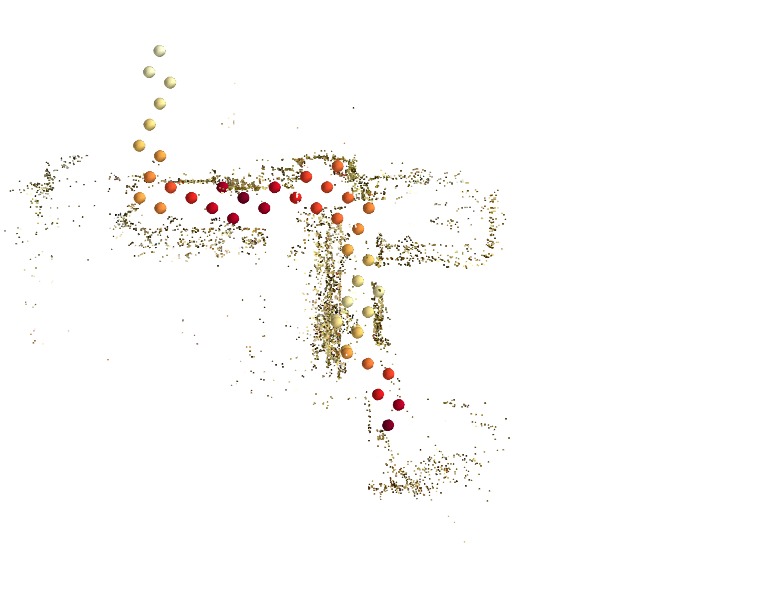}
\end{subfigure}
\begin{subfigure}[c]{.33\textwidth}
\centering
\includegraphics[width=\textwidth]{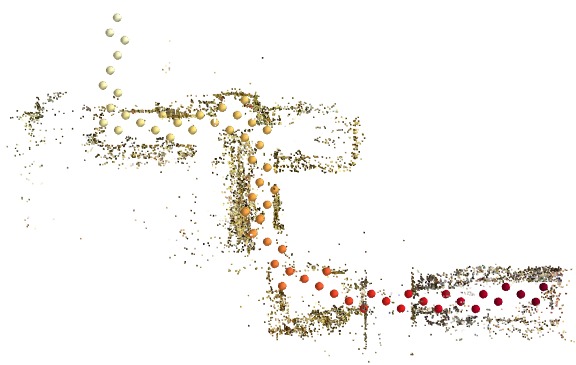}
\end{subfigure}
\begin{subfigure}[c]{.33\textwidth}
\centering
\includegraphics[width=\textwidth]{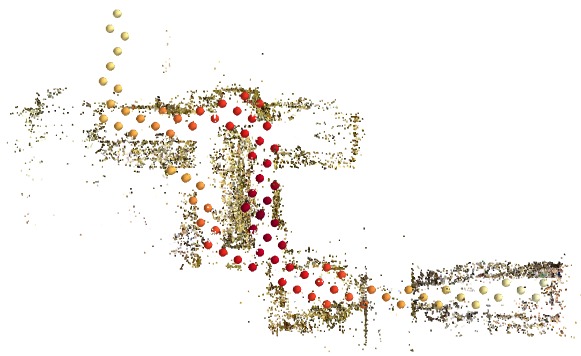}
\end{subfigure}
\begin{subfigure}[c]{.25\textwidth}
\centering
\includegraphics[width=\textwidth]{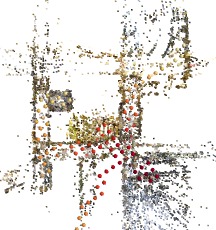}
\end{subfigure}
\begin{subfigure}[c]{.25\textwidth}
\centering
\includegraphics[width=\textwidth]{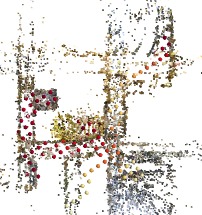}
\end{subfigure}
\begin{subfigure}[c]{.25\textwidth}
\centering
\includegraphics[width=\textwidth]{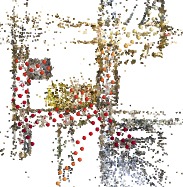}
\end{subfigure}
\caption{Projections of the value function ($V(s)$) for environments as time elapses (left to right). The state space expands as the user visits more locations. For each position, the maximum value (across all states at that position) is displayed:  $\max_{s \in \mathcal{S}_x}{V(s)}$. From top to bottom, the environments are Home 1, Office 1, Lab 1.} \label{fig:value}
\end{figure}

\subsection{RNN baseline settings}
We experimented with a variety of settings for the RNN baseline. After each goal is detected, the RNN is refit. The settings we experimented with are cell $\in$ \{GRU, Basic\}, learning rate $\in \{0.1, 0.01, 0.001, 0.0001\}$, hidden dimension $\in \{8,16,32,64\}$, epochs after each goal $\in \{5,10,50,100\}$.

\fi

\end{document}